\documentclass[10pt,a4paper]{article}
\usepackage[round,authoryear]{natbib}
\usepackage{graphicx}
\usepackage{booktabs}

\usepackage{amsmath,amsfonts,bm}

\def\eqref#1{equation~\ref{#1}}
\def\1{\bm{1}}

\DeclareMathAlphabet{\mathsfit}{\encodingdefault}{\sfdefault}{m}{sl}
\SetMathAlphabet{\mathsfit}{bold}{\encodingdefault}{\sfdefault}{bx}{n}

\usepackage{hyperref}
\usepackage{url}
\usepackage{multirow}

\usepackage{newpxtext}
\usepackage{latexsym}
\usepackage{pifont}
\usepackage{subcaption}
\usepackage{amsmath}

\usepackage{algorithm}
\usepackage{algpseudocode}
\usepackage{enumitem}

\usepackage[T1]{fontenc}
\usepackage[utf8]{inputenc}

\usepackage{microtype}
\usepackage[table]{xcolor}
\usepackage{multirow}
\usepackage[most]{tcolorbox}
\usepackage{booktabs}
\usepackage{longtable}
\usepackage{makecell,tabularx}
\usepackage{listings}
\usepackage{fvextra}
\usepackage{inconsolata}
\usepackage{amssymb}
\usepackage{url}

\let\cite\citep

\algnewcommand{\NoNumberState}[1]{%
  \State\hskip-\ALG@thistlm #1}

\usepackage[table]{xcolor}
\definecolor{OpenBest}{RGB}{226,245,230}
\definecolor{ClosedBest}{RGB}{226,235,252}

\newcolumntype{C}[1]{>{\centering\arraybackslash}m{#1}}

\definecolor{BlueLight}{RGB}{226,235,252}
\definecolor{BlueMain}{RGB}{47,99,171}

\usepackage[left=23mm,right=23mm,top=23mm,bottom=24mm,headheight=15pt,headsep=8mm]{geometry}
\usepackage[scaled=0.92]{helvet}
\usepackage{titlesec,fancyhdr,placeins,needspace}
\definecolor{Ink}{HTML}{122B46}
\definecolor{Teal}{HTML}{008C95}
\definecolor{Mist}{HTML}{EDF6F6}
\definecolor{Gold}{HTML}{D4A34F}
\definecolor{Muted}{HTML}{586A7A}
\hypersetup{colorlinks=true,linkcolor=Teal,citecolor=Teal,urlcolor=Teal,
 pdftitle={Learning to Optimize through Solver-Grounded Self-Play},
 pdfauthor={Xia Jiang, Yaoxin Wu, Chenyu Zhou, Mengzhu Xu, Wim P.M. Nuijten, Yingqian Zhang},bookmarksnumbered=true}
\titleformat{\section}{\Large\sffamily\bfseries\color{Ink}}
 {\colorbox{Teal}{\makebox[1.45em]{\color{white}\strut\thesection}}}{0.7em}{}[\vspace{2pt}{\color{Teal!35}\titlerule}]
\titleformat{\subsection}{\large\sffamily\bfseries\color{Ink}}{\textcolor{Teal}{\thesubsection}}{0.7em}{}
\titleformat{\subsubsection}{\normalsize\sffamily\bfseries\color{Ink}}{\thesubsubsection}{0.7em}{}
\titleformat{\paragraph}[runin]{\normalsize\bfseries\color{Ink}}{}{0pt}{}
\titlespacing*{\section}{0pt}{18pt plus 3pt minus 2pt}{10pt}
\titlespacing*{\subsection}{0pt}{13pt plus 2pt minus 2pt}{6pt}
\titlespacing*{\paragraph}{0pt}{8pt}{0.6em}
\setlist{topsep=4pt,itemsep=2pt,parsep=0pt}
\fvset{fontsize=\small}
\tcbset{fonttitle=\sffamily\bfseries}
\renewcommand{\headrulewidth}{0.4pt}
\renewcommand{\footrulewidth}{0pt}

\begin{document}
\begin{titlepage}
\begin{tikzpicture}[remember picture,overlay]
 \fill[Ink] (current page.north west) rectangle ([yshift=-13mm]current page.north east);
 \fill[Teal] ([yshift=-13mm]current page.north west) rectangle ([yshift=-15mm]current page.north east);
 \draw[Teal!13,line width=1.1pt] ([xshift=-19mm,yshift=-44mm]current page.north east) circle (23mm);
 \draw[Gold!35,line width=1pt] ([xshift=-14mm,yshift=-40mm]current page.north east) circle (17mm);
 \fill[Teal] ([xshift=23mm,yshift=19mm]current page.south west) rectangle ([xshift=65mm,yshift=20mm]current page.south west);
\end{tikzpicture}\par
\vspace*{10mm}
{\sffamily\bfseries\fontsize{31}{37}\selectfont\color{Ink}
Learning to Optimize\newline
through Solver-Grounded\newline
Self-Play\par}
\vspace{6mm}
{\sffamily\fontsize{10}{15}\selectfont\color{Ink}
Xia Jiang\textsuperscript{1}\enspace\enspace
Yaoxin Wu\textsuperscript{1}\enspace\enspace
Chenyu Zhou\textsuperscript{2}\par
Mengzhu Xu\textsuperscript{1}\enspace\enspace
Wim P.M. Nuijten\textsuperscript{1}\enspace\enspace
Yingqian Zhang\textsuperscript{1}\par}
\vspace{3mm}
{\sffamily\small\color{Muted}
\textsuperscript{1}Eindhoven University of Technology\par
\textsuperscript{2}Shanghai Jiaotong University\par}
\vspace{9mm}
{\color{Gold}\rule{20mm}{1.2pt}}\par
\vspace{4mm}
\begin{tcolorbox}[enhanced,colback=Mist,colframe=Mist,boxrule=0pt,
 borderline west={2pt}{0pt}{Teal},arc=0pt,left=5mm,right=5mm,top=4mm,bottom=4mm]
{\sffamily\bfseries\small\color{Teal} ABSTRACT}\par\medskip
Optimization modeling is central to many decision-making scenarios, but traditionally requires extensive domain expertise. While Large Language Models (LLMs) have shown promise in automating this process, current training paradigms mainly rely on human-annotated or teacher-generated datasets. This dependence introduces a \textit{Generalization Ceiling}, where models overfit to narrow data distributions, and \textit{Capability Anchoring}, where models' reasoning is bounded by annotator proficiency and teacher model capability. In response, we propose \textsc{OPT-Zero}, the first fully self-play training framework for optimization modeling that requires zero external training data. \textsc{OPT-Zero} employs a single LLM in a dual-role closed loop: a Proposer that synthesizes increasingly challenging optimization problems alongside their mathematical formulations and solving code, and a Solver that attempts to resolve the problems given only natural-language problem descriptions. Grounded in execution feedback from external optimization solvers, we alternately train both roles using reinforcement learning. This process fosters an auto-curriculum in which the Proposer and Solver co-evolve: generating harder valid problems by the Proposer seamlessly enhances the structural reasoning ability of the Solver. Extensive results indicate that with zero curated data, \textsc{OPT-Zero} matches state-of-the-art data-dependent methods while exhibiting substantially stronger generalizability, establishing self-play training as a highly scalable paradigm for advancing LLM reasoning in modeling and solving optimization problems.
\end{tcolorbox}
\vfill
\end{titlepage}
\setcounter{page}{2}
\section{Introduction}

\begin{figure}[htb]
    \centering
    \includegraphics[width=0.63\linewidth]{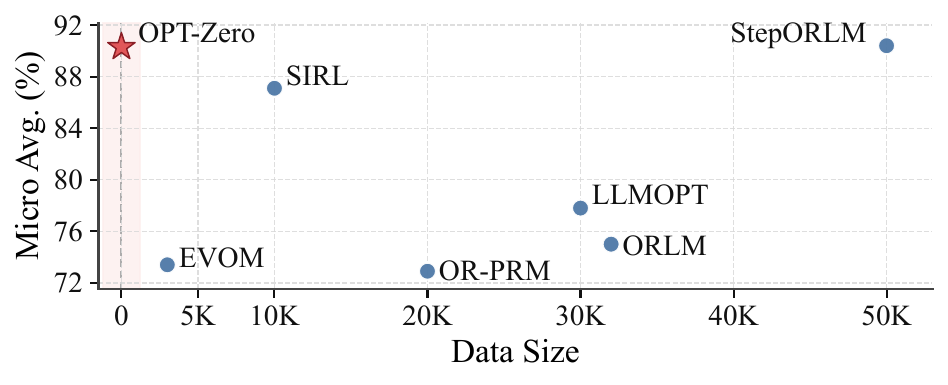}
    \caption{Average performance across six benchmarks vs. annotated training data size.}
    \label{fig:abstract}
\end{figure}

Optimization modeling stands at the core of decision-making across diverse industrial sectors, ranging from supply chain planning \cite{nurjanni2017green, KOCHAKKASHANI2023101602}, healthcare scheduling \cite{earnshaw2003integer, nabavizadeh2024mixed}, to portfolio optimization \cite{hegade2022portfolio, lin2024globally}. Traditionally, this process requires translating complex problems into mathematical formulations and then implementing them in solver-specific languages, which demands significant human effort and expertise \cite{li2025solverllm, le2026making}. Recently, the advancement of Large Language Models (LLMs) has begun to reshape this landscape. Leveraging their language understanding and code generation capabilities, LLMs have enhanced the automation of the modeling process \cite{xiao2023chain, chen2025solverinformed, yazdani2025evocut}. This shift not only streamlines the translation from problem descriptions to formulations, but also makes optimization modeling and solving accessible to a broader range of non-experts \cite{kadiouglu2024ner4opt, jiang2025large}.

LLM-based optimization modeling has technically shifted from inference-time prompt engineering toward training specialized models on optimization-specific data. Initial efforts successfully designed sophisticated workflows and prompt templates to guide LLMs through the modeling process \cite{xiao2023chain, ahmaditeshnizi2024optimus, astorga2025autoformulation, wang2025ormind}. This line of work elicits the latent reasoning potential of general-purpose LLMs by operating within frozen parameters, which may limit the model's ability to deeply internalize specialized optimization knowledge. Consequently, a growing body of recent work focuses on fine-tuning base models with human-crafted, domain-specific datasets to achieve better modeling capabilities \cite{jiang2025llmopt, huang2025orlm, chen2025optimind, lima2025toward, xiao2026deepor,min2026draftandaudit}.

The efficacy of fine-tuning paradigms fundamentally depends on the availability of high-quality datasets \cite{bukharin2024data, zhang2024when}. This dependence is particularly problematic for optimization modeling, where collecting training data requires not only diverse problem descriptions, but also corresponding mathematically valid formulations and solver-verified solutions.
The scarcity of high-quality supervised training examples and the prohibitive overhead of expert annotation constrain the scalability of Supervised Fine-Tuning (SFT) \cite{xiao2025survey}. As illustrated in Table~\ref{tab:comparison}, existing research, ranging from offline SFT frameworks \cite{yang2025optibench} to online Reinforcement Learning (RL) approaches \cite{chen2025solverinformed}, has advanced by leveraging curated training instances that pair natural-language problem descriptions with solver-callable formulations. These instances are typically synthesized with expert intervention \cite{yang2025optibench, jiang2025llmopt} or distilled from teacher models such as GPT-4 \cite{huang2025orlm}.
While establishing strong baselines, their heavy reliance on external training datasets introduces two critical limitations. The first is \emph{Generalization Ceiling}: models trained on human-curated datasets are prone to overfitting specific modeling conventions or solver-dependent syntactic patterns, which undermines their robustness in unseen settings (e.g., when adopting a solver unseen during training) \cite{li2025solverllm}. The second, more fundamental limitation is \emph{Capability Anchoring}: static supervision intrinsically ties the model’s reasoning capability to the proficiency of annotators, either human experts or contemporary teacher LLMs.
As LLMs advance rapidly, existing datasets may lose training value and risk becoming “obsolete” as the reasoning capabilities of LLMs surpass the complexities represented in them. Beyond static dataset curation, LLMs should autonomously explore the optimization landscape, generate problems, and leverage solver feedback as learning guidance. In this regard, a self-evolving training loop is envisioned that continuously expands the model’s competence in optimization modeling.

To remove this dependence on externally curated supervision while retaining reliable solver-grounded feedback, we propose \textsc{OPT-Zero}, a self-play training framework that eliminates the need for human-annotated or teacher-generated data entirely. \textsc{OPT-Zero} operates through a dual-role RL-based training process where a single LLM alternately assumes the roles of Proposer and Solver, engaged in a continuous closed-loop learning. In each iteration, the Proposer generates new optimization problem instances designed to challenge the current capabilities of the Solver. The Solver attempts to formulate models and execute code to solve these instances, receiving feedback from external optimization solvers (e.g., Gurobi, CPLEX). This solver-grounded feedback serves as the reward signal for RL, enabling the LLM to adapt to increasing problem complexity and alternately update the Proposer and Solver policies.
As training progresses, the Proposer learns to synthesize more valid and challenging problems, while the Solver progressively enhances its modeling proficiency to overcome the harder problems. As demonstrated in our experiments, the self-play training creates an auto-curriculum that mitigates issues of generalization ceilings and capability anchoring in static datasets, resulting in competitive performance without relying on annotated data (see Figure \ref{fig:abstract}).

% The self-play training  creates an auto-curriculum that continuously escalates in complexity, mitigating the issues of generalization ceiling and capability anchoring brought by fixed training datasets.

\begin{table}[thb]
\centering
\caption{Comparison of Data Requirements and Training Paradigms. 
\textsuperscript{\dag} Data size with annotated ground-truth code.
\textsuperscript{\ddag} Experts-curated seed data is used.}
\label{tab:comparison}
\renewcommand{\arraystretch}{0.9}
\resizebox{\textwidth}{!}{%
\begin{tabular}{lcccc}
\toprule
\textbf{Method} & \textbf{Data Size}\textsuperscript{\dag} & \textbf{Data Source} & \textbf{Human/Teacher Cost} & \textbf{Paradigm} \\
\midrule
ORLM \cite{huang2025orlm} & 32k & Expert + GPT-4 & 8 Experts + Teacher LLM & SFT \\
LLMOPT \cite{jiang2025llmopt} & $\sim$30k & Expert + GPT-4 & 12 Experts + Teacher LLM & SFT + Alignment \\
OptMATH \cite{lu2025optmath} & 200k & Deepseek-V3 Synthesis & Experts\textsuperscript{\ddag} + Teacher LLM & SFT \\
SIRL \cite{chen2025solverinformed} & 10k & Solver-Assisted Synthesis & Experts\textsuperscript{\ddag} + Teacher LLM & Online RL \\
StepORLM \cite{zhou2026steporlm} & 50k & Expert + GPT-4o & Experts + Teacher LLM & SFT + Alignment \\
OR-PRM \cite{wang2026orprm} & 20k & Expert + GPT-4o & Experts\textsuperscript{\ddag} + Teacher LLM & SFT + Alignment \\
\midrule
\textbf{Ours} & \textbf{0} & \textbf{--} & \textbf{None} & \textbf{Self-Play} \\
\bottomrule
\end{tabular}
}
\end{table}

Our contributions are summarized as follows. 

1) To the authors' knowledge, this is the first study on fine-tuning LLMs for optimization modeling through a fully self-play paradigm. \textsc{OPT-Zero} requires no external training data, eliminating costly expert annotation or teacher LLM distillation, on which previous approaches heavily depend.

2) We develop a solver-grounded RL framework that jointly trains problem generation and solving in a closed loop. \textsc{OPT-Zero} alternates between proposing and solving problems, while external solver execution provides reward signals that guide policy updates, forming a mutually reinforcing pipeline.

3) Without any curated training data,  \textsc{OPT-Zero} can match or even surpass the performance of state-of-the-art data-dependent methods, while showing strong generalizability under different settings, establishing self-play as a scalable training paradigm for advancing LLM in optimization.

% alternative to traditional dataset-based approaches.

\FloatBarrier
\section{Related Work}

\noindent
\textbf{LLMs for Optimization Modeling.} The task of automatically translating natural-language problem descriptions into mathematical programs was formally introduced in the NeurIPS2022 Competition \cite{ramamonjison2023nl4opt}. Since then, there has been a growing interest in leveraging LLMs for optimization and solver-oriented code generation. Existing approaches can be broadly divided into two paradigms: prompt-based and training-based methods. Early prompt-based efforts focused on improving LLM performance through prompt engineering \cite{wasserkrug2024large, chen2024diagnosing}. Subsequent work moved beyond simple prompting to construct more structured pipelines, incorporating strategies such as multi-agent collaboration \cite{xiao2023chain, ahmaditeshnizi2024optimus, deng2024cafa, zhang2025or}, Monte-Carlo tree search \cite{astorga2025autoformulation}, and retrieval-augmented generation \cite{jiang2025droc, liu2025optitree}. Despite improved inference-time reasoning, these methods operate on frozen LLM parameters and thus heavily rely on strong base models. In contrast, training-based methods aim to internalize domain knowledge by fine-tuning on domain-specific datasets \cite{huang2025orlm, jiang2025llmopt, chen2025solverinformed, 11359290, zhou2026steporlm, xiao2026deepor,zhao2026strategyaware}. This paradigm enables small models to acquire specialized modeling skills. However, its effectiveness fundamentally depends on the availability of high-quality data. To alleviate the data bottleneck, recent studies have devoted effort to dataset synthesis pipelines \cite{li2025towards, lu2025optmath, yang2025optibench}. 
Unlike existing paradigms, we eliminate curated supervision from human-annotated or teacher-generated data entirely through self-play, in which LLM modeling capabilities emerge through solver-grounded interactions.

\noindent
\textbf{Self-Play Training in LLMs.} Self-play training generalizes the AlphaZero paradigm \cite{silver2018general} to open-ended domains, improving LLMs through interaction with copies of themselves or their own outputs. \citet{pmlr-v235-chen24j} formalized self-play fine-tuning, showing that a model can bootstrap quality improvements by distinguishing its own previous generations from human-written references, without any new human annotations. Beyond alleviating data scarcity, self-play is also motivated by the availability of cheap automatic verifiers in executable domains (e.g., coding) \cite{haluptzok2023language, wilf2025propose} and by the goal of surpassing static supervised datasets via self-generated curricula \cite{lin2025learning, guo2025deepseek, huang2026rzero}. In reasoning-intensive domains such as mathematics and code generation, self-play has been combined with RL to yield strong results \cite{wu2025selfplay, jayalath2025compute, zhao2025absolute}. 
Our self-play framework differs from prior work as follows:  
1) we focus on optimization modeling and decompose learning into two roles: a Proposer that generates optimization problems and a Solver that formulates and solves them. As a result, the trained model serves dual roles, as both a problem generator and a solver; 2) instead of relying on a fixed task distribution, the Proposer continuously synthesizes new problems. Their difficulty is regulated by a reference buffer with difficulty-aware sampling, producing a self-generated curriculum that co-evolves with the Solver’s capability; 3) both roles are grounded in solver-based verification, which provides RL reward signals and enables the LLM to adapt to increasing problem complexity.

\FloatBarrier
\section{Methodology}

We parameterize a single LLM with weights $\theta$ that is shared between two roles: a Proposer with policy $\pi_\theta^P$ and a Solver with policy $\pi_\theta^S$. Given a context prompt, $\pi_\theta^P$ generates a new optimization problem, comprising a natural-language description, a formal mathematical model (e.g., a linear programming formulation), and executable solver-calling code. After that, $\pi_\theta^S$ receives only the natural-language description of the problem and independently reconstructs the formulation and corresponding code solution.
Under the \textsc{OPT-Zero} framework, the two roles form a mutually reinforcing learning loop. As the Proposer improves, it generates more diverse and challenging problems that expose the Solver to richer scenarios. As the Solver becomes stronger, easy problems provide diminishing training value, which in turn drives the Proposer to produce harder instances.
Crucially, the Proposer is trained to generate valid formulations and code, thus directly enhancing the Solver’s modeling ability. Conversely, by reconstructing formulations from natural language, the Solver training strengthens the model’s structural reasoning, implicitly enabling the Proposer to produce more valid and well-formed problems. This mutual knowledge transfer between two roles, enabled by shared weights $\theta$, is a key advantage of our self-play training framework, eliminating the need for annotators or teacher models.
% over approaches 
% that maintain separate Proposer and Solver networks.

\begin{figure}
    \centering
    \includegraphics[width=0.98\linewidth]{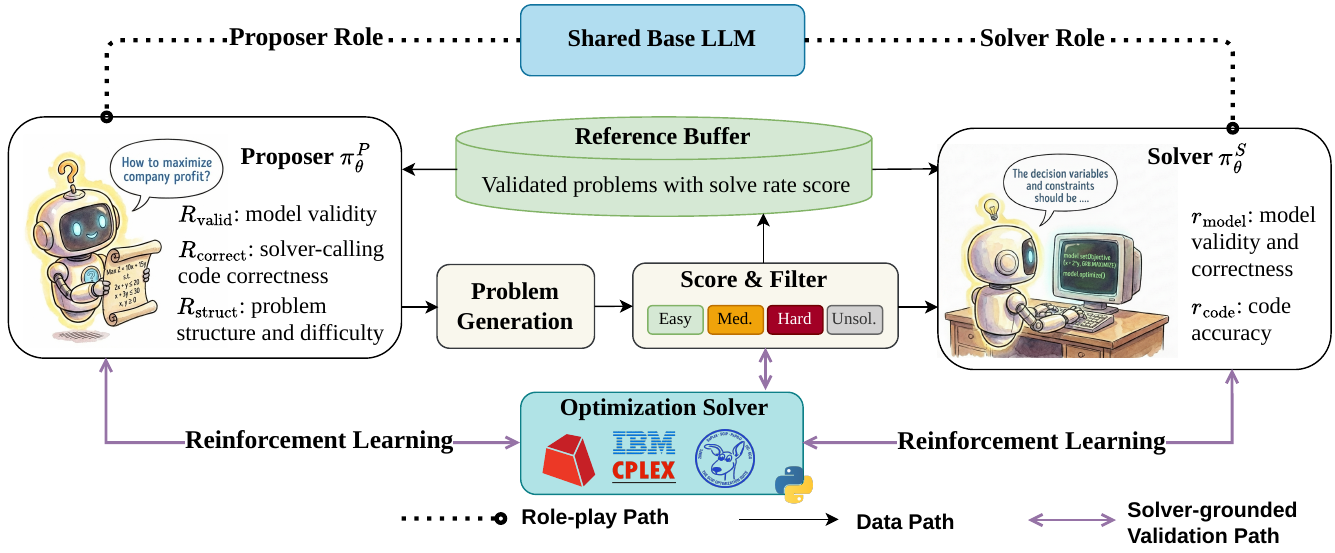}
    \caption{The overall framework of \textsc{OPT-Zero}.}
    \label{fig:framework}
\end{figure}

The training process alternates between optimizing $\pi_\theta^P$ and $\pi_\theta^S$ via Group Relative Policy Optimization (GRPO) over $T$ iterations. Figure~\ref{fig:framework} illustrates the framework with the full loop detailed in Section~\ref{sec:loop}.

\subsection{Problem Formulation}
\label{sec:formulation}

\noindent\textbf{Optimization Problem Representation.}
We consider Linear, Integer, and Mixed-integer Linear Programs (LP/IP/MILP) as the target problem classes in line with the mainstream literature \cite{xiao2023chain, jiang2025llmopt, zhou2026steporlm}. 
% However, \textsc{OPT-Zero} can also be naturally extended to other classes such as Quadratic Programming (QP). 
Formally, a problem instance is represented as a tuple $\mathcal{P} = (\sigma, f, \mathcal{V}, \mathcal{C}, s)$, where $\sigma \in \{\min, \max\}$ is the optimization sense, $f$ is the objective expression over a set of decision variables $\mathcal{V}$, $\mathcal{C}$ is a set of constraints, and $s$ is a natural-language description that presents $\mathcal{P}$ as a practical decision-making scenario without exposing the solution. We use $\mathcal{P}^* = f^*$ to denote the optimal objective for $\mathcal{P}$, which is often obtained by a solver (e.g., Gurobi).

\noindent\textbf{Proposer Output.}
Given a prompt with in-context reference problems sampled from the reference buffer $\mathcal{B}$, the Proposer ($\pi_\theta^P$) is trained to generate a structured completion consisting of the problem tuple $\mathcal{P}$ and the corresponding executable solver-calling code that can produce $\mathcal{P}^*$. Therefore, the Proposer is required to not only generate problems but also provide their solutions in executable code.

\noindent\textbf{Solver Output.}
The solver ($\pi_\theta^S$) receives only the natural-language description $s$ of a problem $\mathcal{P}$ and must independently produce a two-part output: a reconstructed mathematical model, which is defined by $(\sigma, f, \mathcal{V}, \mathcal{C})$, and an executable solver-calling code block. Crucially, $\pi_\theta^S$ has no access to Proposer's formulation (i.e., mathematical model) during generation. The code-level correctness is assessed by comparing the objective value returned by the executed code against $\mathcal{P}^*$, while the declared mathematical model is verified separately for the model-level reward (as detailed below).

\noindent\textbf{Solver-Grounded Verification.}
A central challenge in self-play is obtaining a verified solution, against which the LLM (both Proposer and Solver) outputs can be evaluated. Since the two roles generate mathematical model and code during training, we establish verification processes for optimization modeling and coding, respectively: 1) For optimization modeling, we employ a canonical verifier that independently constructs and solves the mathematical model from the generated model fields $(\sigma, f, \mathcal{V}, \mathcal{C})$ using a canonical solver (e.g., Gurobi). The resulting optimal value $\mathcal{P}^* = f^*$, if exists, serves as the ground-truth objective for the given problem specified by natural language (i.e., $s$). 2) The coding is evaluated separately by executing the generated solver-calling code block in an isolated subprocess, extracting the objective value, and comparing it to $\mathcal{P}^*$. Additional implementation details are provided in Appendix~\ref{app:verification}.

Our goal is to find $\theta$ that simultaneously maximize the quality of generated problems (e.g., validity and structural complexity, see Section~\ref{sec:proposer_reward}) and the accuracy of solutions, by alternating RL updates to $\pi_\theta^P$ and $\pi_\theta^S$ over $T$ iterations. We describe the reward functions and training procedure in the following sections, while full prompts for both Proposer and the Solver are provided in Appendix~\ref{app:prompts}.

\subsection{Proposer Reward Specification}
\label{sec:proposer_reward}

At each iteration, the Proposer $\pi_\theta^P$ receives a prompt containing $K$ reference problems sampled from $\mathcal{B}$ as in-context demonstrations, and is trained to generate $\mathcal{P} = (\sigma, f, \mathcal{V}, \mathcal{C}, s)$ and the corresponding code implementation. The reference problems are used to enhance diversity: by varying the sampled references, we discourage the Proposer from collapsing to a small set of problem structures (e.g., a LP with 3 constraints) or limiting itself to specific application domains of $s$ (e.g., manufacturing).

The Proposer reward combines three factors via a gating formulation: $R^P = R_{\mathrm{valid}} \times R_{\mathrm{correct}} \times R_{\mathrm{struct}}$. Here, $R_{\mathrm{valid}}=\mathbb{I}[\mathcal{P}\ \text{is parseable, feasible, bounded, and solved to optimality}]$ is a hard validity gate. 
Given the objective value $\mathcal{P}^*$ obtained from a canonical solver,
$R_{\mathrm{correct}} \in \{\alpha_c,1\}$ checks whether the Proposer's accompanying code correctly reproduces $\mathcal{P}^*$. Thus, invalid formulations receive zero reward, while valid formulations with incorrect code receive only a capped reward $\alpha_c$. For a valid and self-consistent problem, $R_{\mathrm{struct}} \in [\alpha_s,1]$ measures whether the problem is structurally informative for Solver training. It is computed by the weighted combination of five scores, which capture active variable usage ($s_{\mathrm{var}}$), non-trivial constraint count ($s_{\mathrm{con}}$), coupling density ($s_{\mathrm{den}}$), variable coverage across constraints ($s_{\mathrm{cov}}$), and connectivity among constraints induced by shared variables ($s_{\mathrm{gc}}$). These terms prioritize meaningful optimization structure over superficial size by penalizing duplicated components while rewarding coupled constraints that require relational reasoning among decision variables. The constant $\alpha_s$ is a structural floor that prevents self-consistent problems from receiving a near-zero reward solely because they are simple early in training.

$R_{\mathrm{struct}}$ is not tied to a fixed notion of difficulty. After each iteration, newly generated problems are pre-scored by Solver rollouts, and medium-difficulty problems are used as an empirical estimate of the Solver's current capability frontier (see Section~\ref{sec:loop}). We update the structural target of $R_{\mathrm{struct}}$ from this frontier, so the Proposer is rewarded for generating problems that remain informative as the Solver improves, rather than merely increasing instance size or difficulty in an uncontrolled way. Full definitions of the Proposer reward and its update rule are deferred to Appendix~\ref{app:structural_reward}. We validate these design choices empirically. Section~\ref{sec:ablation} shows that replacing the structured Proposer reward with a binary reward degrades performance, and Appendix~\ref{app:dpo} shows that structural-reward GRPO outperforms a solve-rate-based baseline. These ablations suggest that optimization-specific, fine-grained structural feedback is more effective than coarse validity or solve-rate supervision alone.

\subsection{Solver Reward Specification}
\label{sec:solver_reward}

The Solver $\pi_\theta^S$ receives only the natural-language description $s$ and must independently produce both a formal mathematical model and executable solver-calling code, without access to the Proposer's formulation. The reward $R^S$ is a weighted combination of two components, with $w_m$ and $w_c$ serving as weighting parameters, and is evaluated through independent verification pathways:
\begin{equation}
R^S = w_m r_{\mathrm{model}} + w_c r_{\mathrm{code}}
\label{eq:solver_reward}
\end{equation}
\textbf{Model score} ($r_{\mathrm{model}}$). We compare the objective $f$ of Solver's declared model $(\sigma, f, \mathcal{V}, \mathcal{C})$ against $\mathcal{P}^*$ obtained via our verification pipeline. To provide within-group variance for GRPO, we adopt a rule-based partial-credit scheme. Invalid or unparseable model outputs receive zero reward. Otherwise, the declared model is evaluated through a sequence of validation checks, including correctness of optimization sense, completeness of the objective and constraints, presence of explicit declared variables, dependence of the objective on decision variables, and successful canonical solving. Earlier failures receive smaller fixed rewards, with partial-credit values in $\{0.05, 0.10, 0.15, 0.20\}$ depending on the stage of failure. 
If the declared model passes all checks and is successfully solved, we further evaluate its objective correctness. The reward is defined below with an objective-match indicator $\nu$:
\begin{equation}
r_{\mathrm{model}} = w_{\text{obj}}\times \nu + (1-w_{\text{obj}}), \qquad \nu(y,\mathcal{P}^*) =
\mathbb{I}\!\left[
|y-\mathcal{P}^*| \le
\max(\epsilon_a, \epsilon_r |\mathcal{P}^*|)
\right],
\label{eq:model_match_indicator}
\end{equation}
where $w_{\text{obj}}$ controls the importance of objective matching, and $\nu=1$ when the declared model's objective value $y$ matches $\mathcal{P}^*$. We use $\epsilon_r=10^{-2}$ and $\epsilon_a=10^{-5}$ during training-time verification.

\textbf{Code score} ($r_{\mathrm{code}}$). The Solver's generated code is executed in a sandboxed subprocess under a fixed time limit (i.e., 60$s$). We assign a partial credit based on code execution outcome:
\begin{equation}
    r_{\mathrm{code}} = \begin{cases} 
    w_{\text{obj}} \times \nu + (1-w_{\text{obj}}) & \text{execution succeeds and an objective value is extracted,} \\ 
0.1 & \text{code is present but fails to execute,} 
\\ 0 & \text{otherwise.} \end{cases}
\end{equation}
A floor reward of $0.1$ for failed execution ensures that the LLM is never discouraged from attempting code generation. Again, $r_{\mathrm{model}}$ and $r_{\mathrm{code}}$ are evaluated through independent verification pathways (i.e., model formulation check and code execution check, respectively) so that the Solver must develop consistent reasoning for both model and code formulation rather than overfitting to either one alone.

\subsection{Self-Play Training Loop}
\label{sec:loop}

Both the Proposer and the Solver are optimized with the same GRPO objective
\citep{guo2025deepseek}, augmented with the clip-higher strategy, dynamic
sampling, and token-level loss normalization from DAPO \citep{yu2025dapo}.
The two roles differ only in prompt and reward definition:
for the Proposer, a completion corresponds to a newly proposed optimization
problem (with model formulation and code) and is scored by the proposer-specific reward $R^P$; for the Solver, a
completion corresponds to candidate executable code to a given optimization problem and is scored by the solver-specific reward $R^S$. Concretely, given a prompt $q$, we sample a group of $G$ completions
$\{o_i\}_{i=1}^{G}$ from the current policy and update it with:
\begin{equation}
\mathcal{J}(\theta) = \mathbb{E}_{q}\left[
  \frac{1}{\sum_{i=1}^{G} |o_i|}
  \sum_{i=1}^{G} \sum_{t=1}^{|o_i|}
  \min\left(
    \rho_{i,t}\hat{A}_i,\;
    \mathrm{clip}\!\left(\rho_{i,t},\, 1-\epsilon,\, 1+\bar{\epsilon}_i\right)\hat{A}_i
  \right)
\right]
-\beta D_{\mathrm{KL}}\!\left[\pi_\theta \,\|\, \pi_{\mathrm{ref}}\right],
\label{eq:grpo}
\end{equation}
where $t$ indexes tokens within $o_i$; $\rho_{i,t}=\frac{\pi_\theta(o_{i,t}\mid q,o_{i,<t})}
{\pi_{\mathrm{old}}(o_{i,t}\mid q,o_{i,<t})}$ is the token-level importance
ratio between the current policy $\pi_\theta$ and the sampling policy $\pi_{\mathrm{old}}$, where $\pi_{\mathrm{old}}$ denotes the policy checkpoint at the beginning of the current training iteration;
$R_i$ is the reward of $o_i$, and $\hat{A}_i = (R_i-\mu_r)/\sigma_r$ is the group-normalized advantage computed
from rewards of the $G$ completions.
$\mu_r$ and $\sigma_r$ denote the within-group mean and standard deviation,
respectively; $\epsilon$ is the base clipping coefficient; and $\bar{\epsilon}_i=\epsilon_h\mathbf{1}[\hat{A}_i>0]
+\epsilon\mathbf{1}[\hat{A}_i\leq 0]$ is the asymmetric upper clipping bound
with $\epsilon_h>\epsilon$. Moreover, the KL (Kullback-Leibler) penalty with coefficient $\beta$ constrains updates relative to policy $\pi_{\mathrm{ref}}$.

Since both problem proposal and problem solving exhibit sparse and highly prompt-dependent rewards, using group-relative advantages removes prompt-level reward scale variation and makes training depend on the relative quality of completions within the same context. Normalizing by the total number of generated tokens stabilizes the overall update scale across groups with different completion lengths, which is important because both Proposer and Solver generations vary substantially in length. The clip-higher modification allows larger policy improvements on high-reward completions while keeping conservative updates on weaker ones, which is especially helpful during early training when successful trajectories are rare. As in DAPO, we discard groups whose $G$ completions receive identical rewards, since such groups yield zero relative advantage and provide no learning signal. At each iteration $t$, training proceeds in four stages, as presented by Algorithm~\ref{alg:optzer} in Appendix~\ref{app:algorithm}.

\noindent
\textbf{Stage 1: Proposer Training.}
We construct a proposer training set $\mathcal{D}_t^P$ of $N_P$ examples, each consisting of a prompt that includes $K$ reference problems sampled from the current buffer $\mathcal{B}_{t-1}$ to encourage diversity ($K=0$ in the first iteration). The Proposer ($\pi^P_{\theta_{t-1}}$) is then updated with GRPO
on $\mathcal{D}_t^P$ using reward $R^P$, producing intermediate
parameters $\tilde{\theta}_t$, where $R^P$ is defined in Section~\ref{sec:proposer_reward}.

\noindent
\textbf{Stage 2: Proposer Generation and Pre-scoring.}
The updated Proposer $\pi^P_{\tilde{\theta}_t}$ generates
$\hat{N}$ candidate problems with objective values obtained
through the solver-grounded verification process described
in Section~\ref{sec:formulation}. Each valid candidate $p$ is
pre-scored using $H$ independent rollouts from the Solver.
Reusing the objective-matching indicator $\nu$ defined in
Eq.~\ref{eq:model_match_indicator}, 
we define:
\begin{equation}
\hat{\rho}(p) = \frac{1}{H}\sum_{h=1}^{H}\nu(y_h,\mathcal{P}^*),
\end{equation}
where $y_h$ is the objective value returned by the $h$-th Solver rollout; failed executions are assigned $\nu(y_h,\mathcal{P}^*)=0$. Problems are then classified into difficulty categories based on solve rate $\hat{\rho}$:
\begin{equation}
d(p) =
\begin{cases}
\text{hard},   & 0 < \hat{\rho}(p) \leq 0.25 \\
\text{medium}, & 0.25 < \hat{\rho}(p) \leq 0.75 \\
\text{easy},   & 0.75 < \hat{\rho}(p) < 1
\end{cases}
\end{equation}

\noindent
\textbf{Stage 3: Buffer Update.}
After pre-scoring, we retain only problems that are valid and informative for subsequent learning. In particular, we discard problems with $\hat{\rho}(p)=0$, which provides no reliable training signal, as well as trivial problems with $\hat{\rho}(p)=1$, which are already fully mastered by the current Solver. The remaining problems are admitted into the reference buffer $\mathcal{B}_t$, a capped repository of previously generated instances used for both Proposer prompting and Solver training. To prevent learning collapse toward a narrow set of templates, we impose a structural deduplication that limits the number of problems sharing the same signature, including the problem family, number of variables, number of constraints, and optimization sense. When the buffer exceeds its maximum capacity $|\mathcal{B}|_{\max}$, entries are evicted based on their estimated training value, with easier instances removed before medium ones and medium instances removed before harder ones, ensuring that $\mathcal{B}$ remains concentrated on diverse and sufficiently challenging problems.

% with low-utility problems removed in priority over more challenging ones. Concretely, easy instances are removed before medium ones, and medium instances before hard ones, so that $\mathcal{B}_t$ remains concentrated on diverse and sufficiently challenging problems.

\noindent
\textbf{Stage 4: Solver Training and Difficulty Refresh.}
We construct the Solver training set $\mathcal{D}_t^S$ from two
complementary sources: newly generated problems retained after
pre-scoring and problems sampled from
the reference buffer $\mathcal{B}_t$. The new problems make
Solver updates responsive to the evolving Proposer, while
buffer replay helps mitigate short-term distribution drift
and maintain coverage of previously discovered problem families. Specifically, 80\% of the samples in $\mathcal{D}_t^S$ are drawn
from the current iteration's retained problems, while the
remaining 20\% are sampled from $\mathcal{B}_t$. Both sources
are sampled according to the fixed difficulty-aware distribution
$(w_{\text{easy}}, w_{\text{med}}, w_{\text{hard}})
= (0.2, 0.3, 0.5)$, which prioritizes harder instances.
Starting from the intermediate parameters $\tilde{\theta}_t$,
the Solver is updated with GRPO
on $\mathcal{D}_t^S$ using reward $R^S$, producing the
end-of-iteration parameters $\theta_t$. After Solver training, we refresh the empirical solve-rate
estimates and difficulty labels of replayed buffer problems
using the rollouts collected during this training stage. These rollout-based estimates provide an online approximation of problem difficulty relative to the evolving Solver.

\FloatBarrier
\section{Experiments}

We instantiate \textsc{OPT-Zero} from \textsc{Qwen3-8B-Base} and optimize it with GRPO using the \texttt{ms-swift}. We run 15 self-play iterations, using 64 Proposer prompts and 64 Solver training samples per iteration. We evaluate on 6 datasets: NL4Opt \cite{ramamonjison2023nl4opt}, MAMO-EasyLP \cite{huang2024mamo}, MAMO-ComplexLP \cite{huang2024mamo}, NLP4LP \cite{ahmaditeshnizi2024optimus}, IndustryOR (IndOR) \cite{huang2025orlm}, and ReSocratic \cite{yang2025optibench}, which are widely used benchmarks, with details provided in prior literature \cite{xiao2025survey, zhou2026steporlm}. We report Pass@1 accuracy, defined as the proportion of test instances for which a single generated program returns the correct objective value. We report per-dataset results and the macro- and micro-average metrics across all benchmarks. More details of experiment settings are provided in Appendix~\ref{app:setting}.

We consider three types of baselines, including zero-shot LLMs, inference-time methods, and fine-tuned LLMs. Zero-shot LLMs include OpenAI o4-mini \cite{openai_o4mini_2025}, DeepSeek-R1 \cite{guo2025deepseek}, GLM-5 \cite{zeng2026glm}, Google Gemma4-31B, Qwen3-14B and Qwen3-8B \cite{yang2025qwen3}. Inference-time methods (all based on GPT-4o) include OptiMUS \cite{ahmaditeshnizi2024optimus}, Chain-of-Thought (CoT) strategy \cite{wei2022chain}, Chain-of-Experts (CoE) \cite{xiao2023chain}, and CAFA \cite{deng2024cafa}. For fine-tuned LLMs, we compare against ORLM \cite{huang2025orlm}, LLMOPT \cite{jiang2025llmopt}, OR-PRM \cite{wang2026orprm}, EVOM \cite{guan2026execution}, SIRL \cite{chen2025solverinformed}, and StepORLM \cite{zhou2026steporlm}. Since  \textsc{OPT-Zero} can be extended by introducing existing dataset during Solver training, we also report results of this enhanced version (\textsc{OPT-Zero}-D), which samples 30\% problems per iteration from training set of \citet{chen2025solverinformed}.

\begin{table}[t]
\footnotesize
\centering
\caption{The overall performance of \textsc{OPT-Zero} and baselines with Pass@1 accuracy (\%). *: results of the model are drawn from the original literature.}
\label{table:main_results}
\setlength{\tabcolsep}{2pt}
\renewcommand{\arraystretch}{0.86}
\begin{tabularx}{\textwidth}{l*{8}{>{\centering\arraybackslash}X}}
\toprule
Model & NL4Opt & EasyLP & ComplexLP & NLP4LP & IndOR & ReSocratic & Macro Avg. & Micro Avg. \\
\midrule
\multicolumn{9}{c}{\bf \it Zero-shot LLMs} \\
\textsc{o4-mini} & 78.9 & 90.5 & 56.8 & 87.6 & 69.0 & 74.9 & 76.3 & 81.2 \\
\textsc{DeepSeek-R1} & 78.9 & 89.5 & 59.5 & 87.6 & 59.5 & 84.6 & 76.6 & 83.4 \\
\textsc{GLM-5} & 85.0 & 91.4 & 70.3 & 89.9 & 69.1 & 83.1 & 81.5 & 85.9 \\
\textsc{Gemma4-31B} & 87.3 & 89.4 & 66.7 & 87.6 & 69.0 & 83.9 & 80.7 & 85.1 \\
\textsc{Qwen3-14B} & 70.4 & 87.5 & 45.0 & 79.8 & 54.8 & 73.0 & 68.4 & 76.1 \\
\textsc{Qwen3-8B} & 70.0 & 86.8 & 21.6 & 68.5 & 42.9 & 69.5 & 59.9 & 71.5 \\
\midrule
\multicolumn{9}{c}{\bf \it Inference-time Methods} \\
\textsc{OptiMUS-v0.3} & 76.2 & 78.0 & 46.8 & 88.8 & 45.2 & 87.6 & 70.4 & 78.4 \\
\textsc{CoT} & 62.2 & 49.5 & 42.3 & 74.7 & 40.5 & 43.6 & 52.1 & 51.9 \\
\textsc{CoE} & 66.7 & 94.4 & 50.6 & 87.4 & 31.2 & 71.2 & 66.9 & 78.3 \\
\textsc{CAFA} & 68.1 & 71.2 & 44.5 & 50.0 & 41.1 & 40.1 & 52.5 & 57.0 \\
\midrule
\multicolumn{9}{c}{\bf \it Fine-tuned LLMs} \\
\textsc{ORLM} & 73.8 & 90.4 & 59.5 & 76.4 & 42.9 & 61.8 & 67.5 & 75.0 \\
\textsc{LLMOPT} & 75.1 & 83.5 & 67.6 & 86.0 & 52.4 & 73.2 & 73.0 & 77.8 \\
\textsc{OR-PRM}\textsuperscript{*} & 67.6 & 89.4 & 12.6 & 86.5 & 45.2 & 66.7 & 61.3 & 72.9 \\
\textsc{EVOM}\textsuperscript{*} & 84.1 & 88.2 & 34.3 & - & 31.0 & 63.0 & 60.1 & 73.4 \\
\textsc{SIRL} & 91.5 & 97.8 & 45.9 & 92.7 & 45.2 & 83.4 & 76.1 & 87.1 \\
\textsc{StepORLM} & 96.7 & 97.6 & 77.5 & 97.2 & 52.4 & 81.9 & 83.9 & 90.4 \\

\textsc{OPT-Zero} & 93.0 & 96.7 & 58.6 & 96.6 & 54.8 & 89.8 & 81.6 & 90.3 \\

\textsc{OPT-Zero}-D & 94.4 & 97.6 & 61.3 & 99.4 & 54.8 & 89.8 & 82.9 & 91.3 \\
\bottomrule
\end{tabularx}
\end{table}

\subsection{Main Results}
Table~\ref{table:main_results} compares \textsc{OPT-Zero} with all baselines on six benchmarks. Remarkably, although \textsc{OPT-Zero} is only an 8B model, it outperforms substantially larger general-purpose LLMs such as GLM-5 (744B), as well as prompt-based inference methods built on GPT-4o. Example Proposer generations and Solver solutions of \textsc{OPT-Zero} are provided in Appendix~\ref{app:proposer_gen} and Appendix~\ref{app:example}, respectively.

Even without using any expert-annotated data for fine-tuning, \textsc{OPT-Zero} surpasses most data-dependent approaches and reaches performance comparable to state-of-the-art methods such as StepORLM. In particular, among fine-tuned LLMs, \textsc{OPT-Zero} achieves new state-of-the-art results on IndustryOR (54.8) and ReSocratic (89.8). 
Further finetuned by a small amount of data (only problem description and optimal objectives), the  micro-average performance of \textsc{OPT-Zero}-D  surpasses all previous methods.
Importantly, unlike prior SFT-based methods, these gains do not come at the cost of general-purpose reasoning capabilities (see Appendix~\ref{app:general}). In addition, we also evaluate the generalizability of the trained model to other optimization solvers in Appendix~\ref{app:transfer}, where we find \textsc{OPT-Zero} exhibits strong zero-shot transferability, which existing state-of-the-art methods fail to demonstrate. These results highlight the potential of \textsc{OPT-Zero} in overcoming the issue of generalization ceiling. Finally, Appendix~\ref{app:scaling} shows that \textsc{OPT-Zero} can further benefit from inference-time scaling, leading to additional performance gains beyond the single-sample setting.

\subsection{Ablation Studies}
\label{sec:ablation}

To assess the contribution of each component, we evaluate five variants of \textsc{OPT-Zero}: 1) w/o Solver training; 2) w/o Proposer training; 3) w/o Reference buffer, where no buffer is used in either the Proposer prompts or the Solver training; 4) w. Binary Solver reward, where the structured Solver reward is replaced by a binary signal, assigning 1 for correct generations and 0 for all other outputs; and 5) w. Binary Proposer reward, which applies the same binary scheme to the Proposer.

As shown in Table~\ref{table:ablation}, removing any major component or simplifying the reward consistently hurts performance, confirming the effectiveness of our methodological design. The largest accuracy drops come from removing Solver training or replacing its structured reward with a binary signal, indicating that Solver training is central to solving optimization problems. Proposer training and its structured reward also improve Solver performance, clearly suggesting beneficial mutual interaction between the two roles. The reference buffer is likewise important, as it helps the Proposer generate harder problems while providing useful problem-distribution memory for Solver learning.

In addition, we further investigate the choice of sampling weights in the difficulty-aware Solver training process in Appendix~\ref{app:difficulty}, as well as the use of structural reward in Proposer RL in Appendix~\ref{app:dpo}.

\begin{table*}[t]
\footnotesize
\centering
\caption{Ablation study of \textsc{OPT-Zero} on Pass@1 accuracy (\%).}
\label{table:ablation}
\setlength{\tabcolsep}{3pt}
\renewcommand{\arraystretch}{0.9}
\resizebox{\textwidth}{!}{%
\begin{tabular}{ccccccccc}
\toprule
Model & NL4Opt & EasyLP & ComplexLP & NLP4LP & IndOR & ReSocratic & Macro Avg. & Micro Avg. \\
\midrule

\textsc{OPT-Zero} & 93.0 & 96.7 & 58.6 & 96.6 & 54.8 & 89.8 & 81.6 & 90.3 \\

w/o Solver training & 77.9 & 93.8 & 45.0 & 75.8 & 45.2 & 76.7 & 69.1 & 79.8 \\
w/o Proposer training & 87.8 & 93.6 & 48.7 & 89.3 & 50.0 & 80.9 & 75.1 & 84.3 \\

w/o Reference buffer & 84.5 & 94.0 & 40.5 & 87.1 & 50.0 & 81.4 & 72.9 & 83.2 \\

w. Binary Solver reward & 83.1 & 94.9 & 24.3 & 83.1 & 47.6 & 75.2 & 68.0 & 79.9 \\
w. Binary Proposer reward & 91.5 & 95.8 & 50.5 & 93.8 & 45.2 & 83.6 & 76.7 & 86.9 \\

\bottomrule
\end{tabular}
}
\end{table*}

\begin{figure*}[t]
    \centering
    \begin{minipage}[t]{0.62\textwidth}
        \centering
        \includegraphics[width=\textwidth]{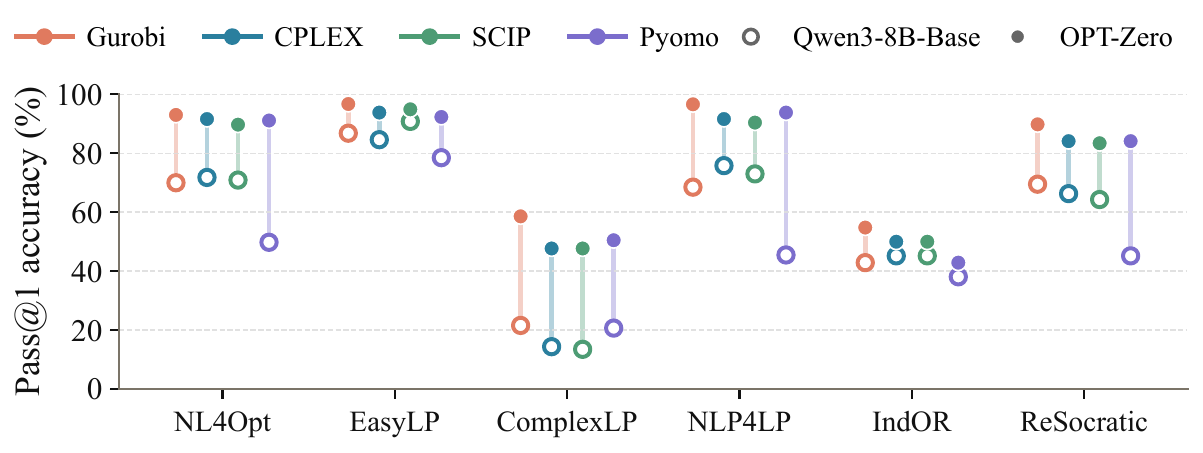}
        \captionsetup{width=0.96\linewidth}
        \captionof{figure}{Accuracy comparison between \textsc{Qwen3-8B-Base} and \textsc{OPT-Zero} across 4 solvers. Filled markers denote \textsc{OPT-Zero}, while hollow markers denote the base model.}
        \label{fig:solver_comparison_dumbbell}
    \end{minipage}
    \hspace{0.01\textwidth}
    \begin{minipage}[t]{0.35\textwidth}
        \centering
        \includegraphics[width=\textwidth]{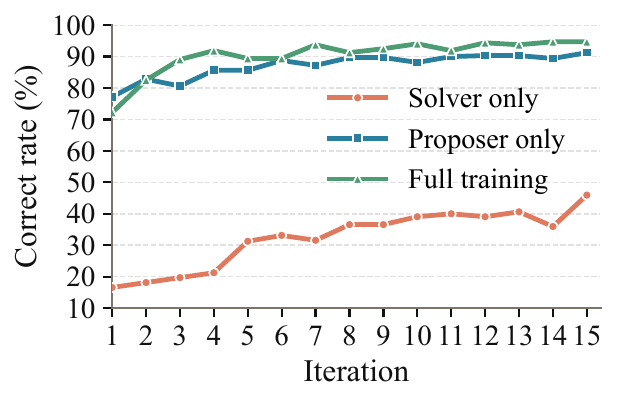}
        \captionsetup{width=0.96\linewidth}
        \captionof{figure}{Proposer generation correct rate over self-play iterations under different training settings.}
        \label{fig:proposer_rate}
    \end{minipage}
\end{figure*}

\subsection{Discussion}

\noindent
\textbf{Learning to Use Different Solvers.} As a generalizable framework, \textsc{OPT-Zero} can be applied to train LLMs to use a variety of solvers and optimization languages. Beyond Gurobi, we train \textsc{Qwen3-8B-Base} to interact with CPLEX via DOcplex, SCIP via PySCIPOpt, and Pyomo. Figure~\ref{fig:solver_comparison_dumbbell} compares the performance of \textsc{Qwen3-8B-Base} and our trained models. According to the figure, while the degree of improvement differs, \textsc{OPT-Zero} consistently enhances model performance across different solvers. More importantly, since these improvements are obtained by training with the target solver, we further verify zero-shot transferability of \textsc{OPT-Zero} across solvers in Appendix~\ref{app:transfer}.

\noindent
\textbf{Synergy between Proposer and Solver.}
The Proposer and Solver are designed to learn cooperatively through shared parameters. To examine \textit{how the Solver benefits the Proposer}, we track the correctness rate (i.e., $R_{\mathrm{correct}}=1$) of the $\hat{N}$ problems generated by the Proposer at each iteration in Figure~\ref{fig:proposer_rate}. Training the Solver alone already improves Proposer correctness, while jointly training both roles achieves the best performance, indicating effective knowledge transfer through shared parameters. Conversely, to examine \textit{how the Proposer benefits the Solver}, Table~\ref{table:ablation} shows that Solver-only training underperforms joint training by a clear margin (75.1 vs. 81.6 Macro Avg.), suggesting that Proposer training supplies more informative problem instances for Solver improvement.

\noindent
\textbf{Generalizability to Different LLMs.} In Appendix~\ref{app:llms}, we demonstrate that the gains of our method persist across LLMs with different architectures and sizes, suggesting that the improvement is not tied to a particular base model. Instead, it provides adaptive supervision that scales with the LLM’s current modeling frontier, mitigating capability anchoring to a static training dataset.
\FloatBarrier
\section{Conclusion}

We present \textsc{OPT-Zero}, a framework that enables LLMs to master optimization tasks without relying on human-annotated data. By iteratively training dual roles (i.e., a Proposer and a Solver) through RL, \textsc{OPT-Zero} establishes a self-play loop that progressively enhances LLMs in optimization reasoning. Extensive experiments demonstrate that \textsc{OPT-Zero} achieves competitive performance among LLM-based frameworks, significantly outperforming the backbone model on diverse benchmark tasks.

Future directions include: 1)  extending the framework to broader problem classes, such as nonlinear programming and stochastic programming; 2) developing more refined reward mechanisms (e.g., training a reward model) to better assess Proposer and Solver outputs; 3) integrating solver-calling code generation with automated heuristic discovery to further amplify optimization performance.
\clearpage
\bibliographystyle{plainnat}
\bibliography{references}
\clearpage
{\sffamily\bfseries\small\color{Teal} SUPPLEMENTARY MATERIAL\par}
\appendix
\appendix

\section{Canonical Verification Pipeline}
\label{app:verification}

Since the Proposer and the Solver are all required to generate both mathematical model and code, the generation verification process involves verifying mathematical model fields and verifying code field, respectively. The overall procedure is detailed as below.

\paragraph{Mathematical Model Verification.} The structured output of the optimization model is parsed with a regex-based extractor that identifies seven tagged fields: \texttt{family}, \texttt{sense}, \texttt{objective}, \texttt{variables}, \texttt{constraints}, \texttt{story}, and \texttt{code}. Variable specifications follow the format \texttt{name:type:lb:ub} (type $\in$ {\texttt{C}, \texttt{I}, \texttt{B}} for continuous, integer, binary), and array-style declarations (e.g., \texttt{x[0:200]:I:0:1}) are expanded element-wise up to a limit of $10^4$ variables. A completion is considered format-valid if all of {\texttt{sense}, \texttt{objective}, \texttt{constraints}, \texttt{story}} are non-empty. Before model construction by the solver, all mathematical expressions undergo a normalization pass: Unicode operators ($-$, $\times$, $\div$, $\leq$, $\geq$) are mapped to ASCII equivalents; variable names are standardized (e.g., \texttt{x\_1} $\to$ \texttt{x1}, \texttt{x[1]} $\to$ \texttt{x1}); implicit multiplication is inserted between adjacent identifiers and numerals (e.g., \texttt{3x1} $\to$ \texttt{3*x1}); and constraint shorthand such as chained inequalities (\texttt{0 <= x <= 10}) and grouped bounds (\texttt{x1, x2 >= 0}) are expanded into atomic comparisons. Expressions are then parsed into Python ASTs and evaluated with degree tracking; objectives with degree $\geq2$ and constraints with degree $>1$ are rejected (because we only consider LP/IP/MILP). The normalized fields are used to programmatically construct a Gurobi model with \texttt{OutputFlag=0} and a configurable wall-clock \texttt{TimeLimit} (60s in our setting). Variables are added with their declared types and bounds; the objective and all constraints are added via the evaluated AST. If Gurobi returns \texttt{OPTIMAL}, the objective value $\mathcal{P}^*$ is recorded; \texttt{INFEASIBLE} or \texttt{UNBOUNDED} outcomes cause the problem to be invalid with zero reward.

\paragraph{Code Verification.}
The generated solver-calling code is written to a temporary file and executed in a subprocess via \texttt{subprocess.run} with a hard timeout (default 60 s). The verifier scans stdout for the pattern \texttt{objective\_value={value}} and parses the floating-point result. Execution is considered successful only if the process exits with code 0 and the pattern is found; otherwise the code component is regarded as invalid. The subprocess isolation ensures that infinite loops, illegal memory accesses, or solver license errors in generated code cannot corrupt the training process. We compare the objective $y$ of the code to $\mathcal{P}^*$ via a relative tolerance criterion:
$\text{correct} = \mathbf{1}\left[\frac{|y - \mathcal{P}^*|}{|\mathcal{P}^*|} \leq \epsilon_r\right],$
where $\epsilon_r$ is a relative tolerance (we use $\epsilon_r = 10^{-2}$; absolute tolerance $\epsilon_a = 10^{-5}$ is used when $|\mathcal{P}^*| < 10^{-9}$).

\section{LLM Prompts}
\label{app:prompts}

We present the prompts for both the Proposer and the Solver as below. For the Proposer, we use Seed Proposer Prompt for the first iteration and use Proposer Prompt for the rest of the training iterations, as there is no reference problems in the buffer $\mathcal{B}$ at the first iteration.

\begin{tcolorbox}[
  enhanced,
  breakable,
  colback=white,
  colframe=black,
  boxrule=0.4pt,
  arc=1.5mm,
  left=1mm,
  right=1mm,
  top=1mm,
  bottom=1mm,
  title=Seed Proposer Prompt
]
\begin{Verbatim}[breaklines=true,breakanywhere=true]
Invent challenging but solvable optimization tasks that yourself can learn from.

- Please randomly choose a model family from <IP (Integer Programming )| LP (Linear Programming) | MILP (Mixed-Integer Linear Programming)>.
- Variable types: C=continuous, I=integer, B=binary. If a variable appears in constraints that imply counting (e.g., sum of selected items, assignment), it should be integer or binary.
- For many similar variables (same type and bounds), you MAY use indexed notation:
  - Variables: x[0:N]:B:0:1 (expands to x0, x1, ..., x(N-1)), where N is your chosen count
  - Objective/constraints: sum(x[i] for i in range(N))
  - Only sum() with range() is supported; no nested loops or conditionals in generator expressions.
- LP/IP/MILP only: objective and constraints must be linear.
- You MUST also provide a correct {SOLVER} implementation of the SAME model (see ###code below).
- In ###objective and ###constraints:
  - Use explicit "*" for multiplication (no implicit multiplication like "3*x1" or "3 x1").
  - Use only ASCII operators: + - * /
  - Use only ASCII comparisons in constraints: <= >= == (no <= or >= symbols, no chained inequalities like "0 <= x1 <= 10").
  - Each constraint must contain exactly ONE comparison; split ranges into two constraints.

OUTPUT (STRICT):
- Return exactly two top-level blocks, in this order:
  1) <thinking> ... </thinking>
  2) <answer> ... </answer>
- No other text outside these tags.
- You MUST include a ###code block containing a complete, executable Python script using {SOLVER_LIB}.
  - The code must match the EXACT ###variables/###objective/###constraints above (same numbers and logic).
  - The code must call optimize() and must print EXACTLY one line: print(f"objective_value={obj:.3f}")
  - Do NOT print anything else. Set solver logs off (e.g., OutputFlag=0).
  - Do NOT wrap code in markdown fences (no ```).
- End sentinel: the final line inside <answer> MUST be exactly "###end". After that, close </answer> and output must end.

<thinking>
Verify feasibility with a concrete assignment, then draft story and code.
</thinking>

<answer>
###family: <IP|LP|MILP>
###sense: <min|max>
###variables: <name:type:lb:ub>; <name:type:lb:ub>; ...
###objective: <math expression, e.g., 3*x1 + 2*x2>
###constraints: <constraints separated by semicolons, e.g., x1 + x2 <= 10; x1 >= 0>
###code:
<Python code using {SOLVER_LIB} that solves the EXACT model above and prints objective_value=...>
###story:
<A real-world story matching the EXACT model above. No "###" or "User:"/"Assistant:" lines.>
###end
</answer>
\end{Verbatim}
\end{tcolorbox}

\begin{tcolorbox}[
  enhanced,
  breakable,
  colback=white,
  colframe=black,
  boxrule=0.4pt,
  arc=1.5mm,
  left=1mm,
  right=1mm,
  top=1mm,
  bottom=1mm,
  title=Proposer Prompt
]

\begin{Verbatim}[breaklines=true,breakanywhere=true]
You will construct a self-contained optimization task that is challenging but solvable for yourself.

Task:
1) Design: choose a model family from <IP (Integer Programming)|LP (Linear Programming)|MILP (Mixed-Integer Linear Programming)>.
2) Formulate: define variables, objective, and constraints mathematically.
3) Verify: ensure feasibility and boundedness.
4) Narrate: write a real-world story that matches the math EXACTLY.

RULES (STRICT):
- LP/IP/MILP only: objective and constraints must be linear.
- You MUST also provide a correct {SOLVER} implementation of the SAME model (see ###code below).
- Story Consistency: EVERY coefficient, constant, and bound in the math MUST appear naturally in the story. No hidden parameters.
- Diversity (model): do NOT propose model with similar objective and constraint as this one: <{Model_REFERENCES_FROM_BUFFER}>. Your model should be more challenging than the given one (e.g., more decision variables, more constraint, complex data format, combinatorial optimization, etc.).
- Diversity (story): do NOT use topics/tones/background similar to this story: <{STORY_REFERENCES_FROM_BUFFER}>. Your story MUST be from a clearly surprising domain and MUST use a different narrative person from the reference story.

OUTPUT (STRICT):
- Return exactly two top-level blocks, in this order:
  1) <thinking> ... </thinking>
  2) <answer> ... </answer>
- No other text outside these tags.
- Formatting: each "###field:" MUST start at the beginning of a new line (no leading spaces).
- Use variable names like x1, x2, y1 (avoid x_1 style if possible).
  Use explicit terms in objective/constraints, e.g., 3*x1 + 2*x2.
- For many similar variables (same type and bounds), use indexed notation instead:
  - Variables: x[0:N]:B:0:1 (expands to x0, x1, ..., x(N-1)), where N is your chosen count
  - Objective/constraints: sum(x[i] for i in range(N))
  - Only sum() with range() is supported; no nested loops or conditionals in generator expressions.
- You MUST include an explicit ###variables block. Variable types: C=continuous, I=integer, B=binary. Bounds use numbers or None.
- If a variable appears in constraints that imply counting (e.g., sum of selected items, assignment), it should be integer or binary.
- In ###objective and ###constraints:
  - Use explicit "*" for multiplication (no implicit multiplication like "3*x1" or "3 x1").
  - Use only ASCII operators: + - * /
  - Use only ASCII comparisons in constraints: <= >= == (no <= or >= symbols, no chained inequalities like "0 <= x1 <= 10").
  - Each constraint must contain exactly ONE comparison; split ranges into two constraints.
- You MUST include a ###code block containing a complete, executable Python script using {SOLVER_LIB}.
  - The code must call optimize() and must print EXACTLY one line: print(f"objective_value={obj:.3f}")
- End sentinel: the final line inside <answer> MUST be exactly "###end". After that, close </answer> and output must end.

<thinking>
Design your model, verify feasibility with a concrete assignment, then draft story and code.
</thinking>

<answer>
###family: <IP|LP|MILP>
###sense: <min|max>
###variables: <name:type:lb:ub>; <name:type:lb:ub>; ...
###objective: <math expression, e.g., 3*x1 + 2*x2>
###constraints: <constraints separated by semicolons, e.g., x1 + x2 <= 10; x1 >= 0>
###code:
<Python code using {SOLVER_LIB} that solves the EXACT model above and prints objective_value=...>
###story:
<A real-world story matching the EXACT model above. No "###" or "User:"/"Assistant:" lines.>
###end
</answer>
\end{Verbatim}
\end{tcolorbox}

\begin{tcolorbox}[
  enhanced,
  breakable,
  colback=white,
  colframe=black,
  boxrule=0.4pt,
  arc=1.5mm,
  left=1mm,
  right=1mm,
  top=1mm,
  bottom=1mm,
  title=Solver Prompt
]
\begin{Verbatim}[breaklines=true,breakanywhere=true]
You will be given a problem description: {PROBLEM_STORY}

Task: extract an optimization model and write a {SOLVER} script using {SOLVER_LIB}.

OUTPUT (STRICT):
- Return exactly two top-level blocks, in this order:
  1) <thinking> ... </thinking>
  2) <answer> ... </answer>
- No other text outside these tags.
- The code must print exactly: print(f"objective_value={obj:.3f}")
- Formatting: each "###field:" MUST start at the beginning of a new line (no leading spaces).
- Use variable names like x1, x2, y1 (avoid x_1 style if possible).
  Use explicit "*" for multiplication (no implicit multiplication like "3x1" or "3 x1").
- For many similar variables (same type and bounds), use indexed notation instead:
  - Variables: x[0:N]:B:0:1 (expands to x0, x1, ..., x(N-1)), where N is your chosen count
  - Objective/constraints: sum(x[i] for i in range(N))
  - Only sum() with range() is supported; no nested loops or conditionals in generator expressions.
- You MUST include an explicit ###variables block. Variable types: I=integer, B=binary, C=continuous. Bounds use numbers or None.
- If a variable appears in constraints that imply counting (e.g., sum of selected items, assignment), it should be integer or binary.
- In ###objective and ###constraints:
  - Use only ASCII operators: + - * /
  - Use only ASCII comparisons in constraints: <= >= == (no chained inequalities like "0 <= x1 <= 10").
  - Each constraint must contain exactly ONE comparison; split ranges into two constraints.
- End sentinel: the final line inside <answer> MUST be exactly "###end". After that, close </answer> and output must end.

<thinking>
Before writing code, do a concrete modeling pass (this is critical):
1) Decision variables (be explicit):
   - List each variable with: name, type (integer/binary/continuous), domain/bounds, and interpretation.
   - Ensure variables are *bounded* (add missing nonnegativity or other bounds implied by the story).
2) Objective (be explicit):
   - State min/max and write the objective formula in terms of your variable names.
3) Constraints (be explicit and labeled):
   - Enumerate constraints one-by-one as algebraic statements (C1, C2, ...).
   - For each constraint, name the variables it uses and any constants.
4) Sanity checks:
   - Boundedness: confirm the objective cannot go to +/- infinity under your constraints/bounds.
   - Feasibility: confirm at least one feasible assignment exists given the story.
5) Implementation plan:
   - Map each variable/constraint/objective to a specific code section.
   - Use the same variable names in code that you used in the model above.
</thinking>


<answer>
###family: <IP|LP|MILP>
###sense: <min|max>
###variables: <name:type:lb:ub>; <name:type:lb:ub>; ...
###objective: <math expression, e.g., 3*x1 + 2*x2>
###constraints: <constraints separated by semicolons, e.g., x1 + x2 <= 10; x1 >= 0>

###code:
{SOLVER_IMPORT}

# 1) Build model
# - Create the model object

# 2) Declare decision variables
# - Use the names from <thinking>
# - Set types and bounds explicitly

# 3) Add constraints
# - Add C1, C2, ... exactly as modeled in <thinking>

# 4) Set objective and optimize
# - Set objective sense (min/max) and expression
# - Optimize and extract objective value as a float `obj`

# Safety: do NOT call quit()/exit()/sys.exit()/os._exit() in generated code.

# CRITICAL: You must print the final objective value exactly like this:
# obj = ...  # a float objective value
# print(f"objective_value={obj:.3f}")
###end
</answer>
\end{Verbatim}
\end{tcolorbox}

\section{Proposer Reward Details}
\label{app:structural_reward}

This section provides the full definition of the Proposer reward introduced in Section~\ref{sec:proposer_reward}. The reward is $R^P = R_{\mathrm{valid}} \times R_{\mathrm{correct}} \times R_{\mathrm{struct}},$ where: the validity gate $R_{\mathrm{valid}} \in \{0,1\}$ checks whether the mathematical model declared by the Proposer can be parsed, constructed, and solved through the solver-grounded verification pipeline. If the model is unparseable, infeasible, or unbounded, then $R_{\mathrm{valid}}=0$ and the final reward is zero; the correctness gate $R_{\mathrm{correct}} \in \{\alpha_c,1\}$ checks whether the Proposer's accompanying code reproduces the canonical objective value $\mathcal{P}^*$. We use a floor value $\alpha_c>0$ for valid formulations whose code is incorrect, while assigning $R_{\mathrm{correct}}=1.0$ when the code executes successfully and matches $\mathcal{P}^*$; 3) the structural term $R_{\mathrm{struct}} \in [\alpha_s,1]$ then scores whether a valid problem is informative for Solver training.

The structural reward is designed to avoid using online solve rate as the Proposer GRPO reward. Directly estimating solve rate during Proposer updates would require multiple Solver rollouts for each sampled Proposer completion, which substantially increases training cost and produces a coarse scalar signal. By contrast, optimization problems expose measurable mathematical structure. We therefore compute $R_{\mathrm{struct}}$ from deterministic properties of the declared formulation, including active variable usage, non-trivial constraint count, coupling density, variable coverage, and connectivity among constraints induced by shared variables. These signals provide dense feedback that discourages structurally degenerate problems, such as duplicated constraints or isolated bounds, while encouraging coupled formulations that require non-trivial modeling reasoning. This structural shaping layer evaluates problem quality through five sub-scores. All sub-scores lie in $[0,1]$ and are computed against adaptive targets $(\bar{v}, \bar{c}, \bar{d})$ that evolve across training iterations. These adaptive targets shift the reward distribution toward the Solver's current capability frontier, as described below.

\paragraph{Saturation function.}
Sub-scores that depend on adaptive targets use a saturation function $\phi(x, \bar{x})$ that peaks at the target and penalizes both under- and over-complexity:
\[
\phi(x, \bar{x}) = \begin{cases}
x / \bar{x} & \text{if } x \leq \bar{x} \\
\exp\left(-0.1\left(\tfrac{x}{\bar{x}} - 1\right)^2\right) & \text{if } x > \bar{x}
\end{cases}
\]
The linear ramp below the target encourages increasing complexity, while the mild Gaussian decay above discourages excessive overshooting without sharply penalizing moderately larger problems (e.g., at $2\times$ the target, $\phi \approx 0.90$, and at $3\times$ the target, $\phi \approx 0.67$).

\paragraph{Sub-score (a): Variable utilization ($s_{\mathrm{var}}$, weight $0.25$).}
Let $\mathcal{V}_{\mathrm{active}} \subseteq \mathcal{V}$ be the set of declared variables that appear either in the objective $f$ or in at least one non-trivial constraint. We define
\[
s_{\mathrm{var}} = \phi(|\mathcal{V}_{\mathrm{active}}|, \bar{v}),
\]
which rewards problems whose number of active variables matches the adaptive target $\bar{v}$.

\paragraph{Sub-score (b): Constraint quality ($s_{\mathrm{con}}$, weight $0.25$).}
Let $n_c$ denote the number of unique non-trivial constraints after expanding chained inequalities and grouped bounds, normalizing each constraint into a canonical textual form, and removing exact duplicates in that normalized form. Let $n_{\mathrm{dup}}$ denote the number of repeated constraints removed by this procedure. We define
\[
s_{\mathrm{con}} = \phi(n_c, \bar{c}) \cdot \left(1 - 0.5 \cdot \frac{n_{\mathrm{dup}}}{\max(1, n_c + n_{\mathrm{dup}})}\right),
\]
where the first factor encourages the number of unique non-trivial constraints to match the adaptive target $\bar{c}$, and the second factor imposes a duplication penalty of up to $50\%$.

\paragraph{Sub-score (c): Coupling density ($s_{\mathrm{den}}$, weight $0.20$).}
Let $\bar{d}_{\mathrm{avg}}$ be the average number of variables per non-trivial constraint. We define
\[
s_{\mathrm{den}} =
\begin{cases}
\min\left(1, \left(\bar{d}_{\mathrm{avg}} / \bar{d}\right)^{0.5}\right) & \text{if } \bar{d}_{\mathrm{avg}} > 1.1, \\
0 & \text{otherwise},
\end{cases}
\]
where $\bar{d}$ is the adaptive density target. The threshold of $1.1$ filters out trivially sparse constraints, such as single-variable bounds.

\paragraph{Sub-score (d): Variable coverage ($s_{\mathrm{cov}}$, weight $0.15$).}
Let $\mathcal{V}_{\mathrm{appear}}$ be the set of declared variables that appear in at least one non-trivial constraint. We define
\[
s_{\mathrm{cov}} =
\frac{\left|\left\{v \in \mathcal{V}_{\mathrm{appear}} : v \text{ occurs in at least two constraints}\right\}\right|}{|\mathcal{V}_{\mathrm{appear}}|}.
\]
In other words, $s_{\mathrm{cov}}$ measures the fraction of appearing variables that participate in more than one constraint. This rewards problems in which variables are shared across multiple constraints, indicating meaningful coupling rather than isolated one-off constraints.

\paragraph{Sub-score (e): Graph connectivity ($s_{\mathrm{gc}}$, weight $0.15$).}
We measure connectivity among constraints induced by shared decision variables. Two constraints are considered connected if they share at least one variable. We define
\[
s_{\mathrm{gc}} = \frac{|\text{largest connected component}|}{n_c},
\]
where the largest connected component is computed over the constraint nodes after linking any pair of constraints that involve a common variable. This penalizes decomposable formulations formed by trivially concatenating independent subproblems.

\paragraph{Unused-variable discount.}
A multiplicative penalty discourages the Proposer from declaring variables that do not participate in the optimization:
\[
\delta_{\mathrm{unused}} = 1 - 0.3 \cdot \frac{|\mathcal{V} \setminus \mathcal{V}_{\mathrm{active}}|}{|\mathcal{V}|}.
\]

\paragraph{Final structural score.}
Combining all components:
\[
R_{\mathrm{struct}} = \max\left(\alpha_s, \bigl(0.25s_{\mathrm{var}} + 0.25s_{\mathrm{con}} + 0.20s_{\mathrm{den}} + 0.15s_{\mathrm{cov}} + 0.15s_{\mathrm{gc}}\bigr) \cdot \delta_{\mathrm{unused}}\right).
\]

\paragraph{Adaptive target updates.}
The targets $(\bar{v}, \bar{c}, \bar{d})$ are initialized by $(5, 5, 2.0)$ and are updated between iterations using the pre-scored problems from the most recent iteration. We first select the subset of problems labeled as medium difficulty, i.e., those with empirical solve rate in $(0.25, 0.75]$. These problems are neither already mastered nor completely unsolved, and therefore provide a stable estimate of the Solver's current capability frontier. If fewer than three such problems are available, the targets are left unchanged. Otherwise, we set the raw target for variable count to the median number of active variables across the selected problems, and the raw target for constraint count to the median number of unique non-trivial constraints. The raw density target is then defined as the ratio between these two medians. The targets are finally smoothed with an exponential moving average with $\alpha_{\mathrm{EMA}} = 0.3$. Hard bounds $\bar{v} \in [4, 80]$, $\bar{c} \in [3, 30]$, and $\bar{d} \in [2.0, 10.0]$ are applied to prevent degenerate targets.

\section{\textsc{OPT-Zero} Self-Play Training Algorithm}
\label{app:algorithm}

\begin{algorithm}[htb]
\caption{\textsc{OPT-Zero} Self-Play Training}
\label{alg:optzer}
\begin{algorithmic}[1]
\Require Initial model $\theta_0$, initial buffer $\mathcal{B}_0=\varnothing$, iterations $T$, proposer prompt count $N_P$, generated candidate count $\hat{N}$, proposer reference count $K$, pre-scoring rollouts $H$
\For{$t = 1, \ldots, T$}

    \Statex \textit{Stage 1: Proposer Training}
    \State Construct proposer prompt set $\mathcal{D}_t^P$ with $N_P$ prompts
    \State For each prompt, sample $K$ reference problems from $\mathcal{B}_{t-1}$ as in-context examples
    \State If $\mathcal{B}_{t-1}=\varnothing$, use the seed proposer prompt with $K=0$
    \State $\tilde{\theta}_t \leftarrow \mathrm{GRPO}(\theta_{t-1}, \mathcal{D}_t^P, R^P)$

    \Statex \textit{Stage 2: Problem Generation and Pre-scoring}
    \State Generate $\hat{N}$ candidate problems $\mathcal{P}_t$ using $\pi_{\tilde{\theta}_t}^P$
    \State Canonically verify each $p \in \mathcal{P}_t$ and discard invalid, infeasible, or unbounded problems
    \For{each valid problem $p$}
        \State Run $H$ Solver rollouts with $\pi_{\tilde{\theta}_t}^S$ on its story $s_p$
        \State Compute empirical solve rate $\hat{\rho}(p)$
        \State Assign difficulty label $d(p) \in \{\text{hard}, \text{medium}, \text{easy}\}$
    \EndFor
    \State Update adaptive structural targets from the pre-scored medium-difficulty problems

    \Statex \textit{Stage 3: Buffer Update}
    \State $\mathcal{P}_t^{\mathrm{keep}} \leftarrow \{p \in \mathcal{P}_t : 0 < \hat{\rho}(p) < 1\}$
    \State $\mathcal{B}_t \leftarrow \mathrm{Admit}(\mathcal{B}_{t-1}, \mathcal{P}_t^{\mathrm{keep}})$
    \State Apply structural deduplication and capacity-based eviction to $\mathcal{B}_t$

    \Statex \textit{Stage 4: Solver Training}
    \State Construct $\mathcal{D}_t^S$ by sampling 80\% from current pre-scored problems and 20\% from $\mathcal{B}_t$
    \State Use difficulty-aware sampling on both sources
    \State $\theta_t \leftarrow \mathrm{GRPO}(\tilde{\theta}_t, \mathcal{D}_t^S, R^S)$
    \State Refresh $\hat{\rho}$ and difficulty labels for replayed buffer problems

\EndFor
\State \Return $\theta_T$
\end{algorithmic}
\end{algorithm}

Algorithm~\ref{alg:optzer} summarizes the full self-play training procedure used in \textsc{OPT-Zero}. Here, $\tilde{\theta}_t$ denotes the intermediate model checkpoint obtained after the Proposer update at iteration $t$, before the subsequent Solver GRPO update. The initial buffer is empty, i.e., $\mathcal{B}_0=\varnothing$. In each iteration, the Proposer is trained with prompts augmented by $K$ reference problems sampled from the current buffer, while each newly generated valid problem is pre-scored using $H$ Solver rollouts to estimate its empirical solve rate $\hat{\rho}(p)$. We retain only informative problems with $0<\hat{\rho}(p)<1$, apply signature-based deduplication and capacity-based eviction when updating the buffer, and then construct the Solver training set by mixing newly generated problems and replayed buffer problems with an 80\%/20\% split under a difficulty-aware sampling.

\section{Detailed Experimental Settings}
\label{app:setting}

Training is conducted on a single server with 4 NVIDIA H100 GPUs and 32 CPU cores in \texttt{bfloat16}. DeepSpeed ZeRO-3 is used to enable a more efficient training. Both roles are trained for one epoch with learning rate $1\times10^{-6}$ and KL coefficient $\beta=0.05$. We adopt the DAPO-style GRPO variant with $\epsilon=0.2$ and $\epsilon_{\text{high}}=0.28$, together with dynamic sampling and token-level loss normalization. We use full-parameter training, a constant learning-rate schedule with 0.05 warmup ratio, per-device batch size 2, gradient accumulation of 4 steps, and gradient checkpointing. During self-play sampling, we set temperature to 1.0, top-$p$ to 0.95, and both maximum sequence length and maximum completion length to 8192 tokens, while sampling 8 completions per prompt group.

The reference buffer stores up to 512 validated problems. For Proposer prompting, we sample $K=1$ reference problem from the current buffer as an in-context exemplar. After generation, each candidate problem is pre-scored with 8 Solver rollouts to estimate its empirical solve rate, and this process is accelerated by the vLLM engine. For the Proposer reward, we set the correctness floor to $\alpha_c=0.3$ and the structural floor to $\alpha_s=0.2$. For the Solver reward, we set $w_m$ to 0.4 and $w_c$ to 0.6. We use $w_{\text{obj}}$=0.7 for objective matching. Solver training uses difficulty-aware sampling with weights $(0.2,0.3,0.5)$ for easy, medium, and hard problems, respectively, and draws samples from newly generated problems and replayed buffer instances with an 80\%/20\% split. To preserve diversity, we cap the number of buffered problems sharing the same structural signature and evict lower-value entries when the buffer exceeds capacity. All generated formulations and code are verified with Gurobi (via \texttt{gurobipy}) under isolated Python execution, with a timeout of 60 seconds per instance.

\paragraph{Benchmark sources and test sets.}
\label{app:evaluation_protocol}
Our six-benchmark evaluation uses the public cleaned suite associated with LLM4OR \cite{xiao2025survey}, also adopted by StepORLM \cite{zhou2026steporlm}. LLM4OR reports inspection by 11 optimization specialists, with each identified error cross-checked by at least three experts. Table~\ref{tab:eval_populations} lists the resulting test subsets, totaling 1,492 problems. The 403 ReSocratic instances are from the cleaned evaluation benchmark associated with OptiBench/ReSocratic \cite{yang2025optibench}. Appendix~\ref{app:challenging} extends the evaluation to OptMATH-Bench and MIPLIB-NL.

\begin{table}[htbp]
\centering
\small
\caption{Cleaned test subsets used in the six-benchmark evaluation.}
\label{tab:eval_populations}
\begin{tabularx}{\textwidth}{l r X}
\toprule
Benchmark & Test instances & Original benchmark reference \\
\midrule
NL4Opt & 213 & \citet{ramamonjison2023nl4opt} \\
MAMO-EasyLP & 545 & \citet{huang2024mamo} \\
MAMO-ComplexLP & 111 & \citet{huang2024mamo} \\
NLP4LP & 178 & \citet{ahmaditeshnizi2024optimus} \\
IndustryOR (IndOR) & 42 & \citet{huang2025orlm} \\
ReSocratic & 403 & \citet{yang2025optibench} \\
\midrule
Total & 1,492 & Public cleaned suite: \citet{xiao2025survey} \\
\bottomrule
\end{tabularx}
\end{table}

\paragraph{Single-generation protocol and correctness criterion.}
For the six-benchmark single-generation evaluations, we generate one solution per problem with temperature 0.1, top-$p$ 0.95, and a maximum of 8192 new tokens. A correct prediction executes successfully within 60 seconds and returns an objective value matching the reference. Following prior optimization-modeling evaluations \cite{liu2025optitree, zhou2026steporlm}, we use a relative tolerance of 5\%, with an absolute tolerance of $10^{-5}$ for near-zero reference objectives. Training-time verification uses a 1\% relative tolerance (Eq.~\ref{eq:model_match_indicator}).

We report objective-value-based Pass@1, which measures end-to-end objective recovery and accommodates equivalent formulations with different variable names or auxiliary variables. The hard-benchmark evaluation in Appendix~\ref{app:challenging} uses the same decoding and scoring configuration. Appendices~\ref{app:scaling} and~\ref{app:coe} evaluate inference-time scaling with multiple candidates and CoE, respectively.

\paragraph{Metric aggregation.}
Let $n_b$ be the size of benchmark $b$ and $c_b$ the number of correct predictions. We report
\begin{equation}
\mathrm{Macro}=\frac{100}{6}\sum_{b=1}^{6}\frac{c_b}{n_b},
\qquad
\mathrm{Micro}=100\frac{\sum_{b=1}^{6}c_b}{1492}.
\label{eq:eval_aggregation}
\end{equation}
Macro-average gives each benchmark equal weight, while Micro-average gives each problem equal weight. On the 42-instance IndustryOR test set, one additional correct prediction increases accuracy by approximately 2.38 percentage points.

\paragraph{Baseline sources and comparisons.}
The zero-shot LLM baselines are accessed through OpenRouter APIs. We re-evaluate publicly accessible optimization fine-tunes where possible; results taken from the original publications are marked with an asterisk in Table~\ref{table:main_results}. Appendix~\ref{app:coe} further compares Qwen3-8B-Base and \textsc{OPT-Zero}-8B using the same CoE inference procedure to examine the benefits of combining self-play training with inference-time scaling.

\subsection{Training Cost and Inference Throughput}
\label{app:compute_cost}

\paragraph{Training-time accounting.}
The main Qwen3-8B self-play run takes approximately 73 hours on four H100 GPUs and 32 CPU cores for 15 iterations, including Proposer training, problem generation and pre-scoring, and Solver training. Table~\ref{tab:training_cost_breakdown} reports the stage times, which sum to 71.4 hours, alongside the approximate end-to-end total.

\begin{table}[htbp]
\centering
\small
\caption{Wall-clock training time for the main Qwen3-8B run on four H100 GPUs.}
\label{tab:training_cost_breakdown}
\begin{tabular}{lr}
\toprule
Component & Wall-clock hours \\
\midrule
Proposer training & 22.9 \\
Problem generation and Solver pre-scoring & 5.2 \\
Solver training & 43.3 \\
\midrule
Itemized subtotal & 71.4 \\
Approximate end-to-end total & 73.0 \\
Difference between total and subtotal (derived) & $\approx 1.6$ \\
\bottomrule
\end{tabular}
\end{table}

\paragraph{Training budget comparison.}
Our Qwen3-8B run uses approximately $4\times73=292$ allocated H100 GPU-hours. For reference, \citet{chen2025solverinformed} report two approximately 24-hour training stages on eight H100 GPUs for SIRL-7B, totaling 384 GPU-hours. Our reported allocation is approximately 24\% lower and includes self-generated curriculum construction. SIRL's reported figure covers its two training stages. \textsc{OPT-Zero} constructs its curriculum through model generation and solver verification, without external problem annotation or teacher-LLM generation calls.

\paragraph{Batched inference at $N=16$.}
For the 1,492-problem suite, generating 16 candidates per problem on one H100 takes 3,813 seconds, and parallel verification takes 172 seconds. The complete workload comprises 23,872 candidates and takes 3,985 seconds, corresponding to an amortized $3985/1492\approx2.67$ seconds per problem including all 16 candidates, or approximately 0.374 problems per second. Generation uses vLLM continuous batching, and verification runs in parallel. Table~\ref{tab:inference_throughput} summarizes these measurements.

\begin{table}[htbp]
\centering
\small
\caption{Timing for the complete $N=16$ evaluation workload, with per-problem time amortized over batched generation and parallel verification.}
\label{tab:inference_throughput}
\begin{tabular}{lr}
\toprule
Quantity & Value \\
\midrule
Test problems & 1,492 \\
Candidates per problem & 16 \\
Total generated candidates & 23,872 \\
Generation hardware & $1\times$ H100 \\
Generation time & 3,813 s \\
Verification time & 172 s \\
Summed generation and verification time & 3,985 s \\
Amortized time per problem, including 16 candidates & $\approx 2.67$ s \\
\bottomrule
\end{tabular}
\end{table}

\section{Additional Results}
\label{app:res}

\subsection{Example Proposer Generations}
\label{app:proposer_gen}

Table~\ref{tab:selected_proposer_stories} shows representative Proposer-generated problems from different self-play iterations (i.e., 0, 4, 9, and 14). We report the iteration index together with the problem family, structural size (Vars refers to the number of variables, while Cons. refers to the number of constraints), and the full natural-language story. We observe a clear qualitative progression over the training process. Early iterations mainly produce small and relatively simple optimization tasks, whereas later iterations yield problems with richer narratives, more variables, and substantially more coupled constraints. This suggests that the Proposer gradually moves from generating straightforward resource-allocation instances to constructing more structured and challenging optimization problems.

Notably, the later examples remain valid and semantically coherent rather than becoming arbitrarily complicated, indicating that the design of our Proposer reward and the self-play mechanism improve both structural sophistication and natural-language problem construction. These qualitative trends are consistent with our quantitative findings and provide further evidence that \textsc{OPT-Zero} induces an effective auto-curriculum: as training proceeds, the Proposer continually expands the complexity frontier of the problems presented to the Solver, thereby supporting sustained capability improvement without external annotated data.

\begin{center}
\small
\setlength{\tabcolsep}{5pt}
\renewcommand{\arraystretch}{1.12}
\begin{longtable}{c c c p{0.73\textwidth}}
\caption{Selected Proposer-generated stories from different self-play iterations. }
\label{tab:selected_proposer_stories} \\
\toprule
Iter. & Vars & Cons. & Full Story \\
\midrule
\endfirsthead

\toprule
Iter.  & Vars & Cons. & Full Story \\
\midrule
\endhead

\bottomrule
\endfoot

0  & 2 & 4 &
A manufacturing company produces two products, A and B. The company has two limited resources: raw materials and labor hours. Product A yields a profit of \$5 per unit and requires 2 units of raw material and 4 labor hours. Product B yields a profit of \$4 per unit and requires 3 units of raw material and 2 labor hours. The company has 120 units of raw material and 100 labor hours available. The goal is to determine the number of units of each product to produce to maximize profit while not exceeding the available resources. \\

4 & 9 & 15 &
A logistics company needs to allocate delivery routes for nine different product categories: Electronics, Clothing, Furniture, Books, Food, Appliances, Toys, Sports Equipment, and Home Decor. Each category has a different profit margin and resource consumption rate. The company has limited resources, including delivery vehicles and staff, which must be managed efficiently to maximize profits.

Electronics require 2 units of delivery resource, Clothing require 3 units, Furniture require 4 units, Books require 1 unit, Food require 2 units, Appliances require 1 unit, Toys require 3 units, Sports Equipment require 2 units, and Home Decor require 1 unit. The company has a total of 100 units of delivery resource.

Additionally, the total number of delivery routes across all categories must not exceed 20 due to staff limitations. There are specific constraints: Electronics and Clothing combined can have at most 10 routes, Furniture and Books combined can have at most 8 routes, Food and Appliances combined can have at most 6 routes, and Toys, Sports Equipment, and Home Decor combined can have at most 12 routes. The profit margins for each category are as follows: Electronics generate a profit of \$50{,}000, Clothing generate \$40{,}000, Furniture generate \$60{,}000, Books generate \$30{,}000, Food generate \$20{,}000, Appliances generate \$50{,}000, Toys generate \$45{,}000, Sports Equipment generate \$35{,}000, and Home Decor generate \$25{,}000. The company wants to determine the optimal number of delivery routes for each product category to maximize total profit while staying within the resource and capacity constraints.
\\

9  & 8 & 13  &
A city planner is tasked with optimizing the allocation of public resources for a major city event. The event has eight candidate resource types, represented by binary variables $x_1,\ldots,x_8$, where $x_i=1$ means that resource type $i$ is selected. Each resource type provides a different benefit: Resource 1 provides 10 units of impact, Resource 2 provides 15, Resource 3 provides 12, Resource 4 provides 8, Resource 5 provides 11, Resource 6 provides 9, Resource 7 provides 7, and Resource 8 provides 14.

The planner must satisfy several operational constraints. At most five resource types can be selected in total. The resource usage coefficients for the eight resource types are $2,3,1,4,2,3,1,$ and $2$, respectively, and the total weighted usage cannot exceed 10 units. In addition, at most two resources can be selected from the first group $\{1,2\}$, at most three from the second group $\{3,4,5,6\}$, and at most two from the third group $\{7,8\}$. The goal is to select the subset of resource types that maximizes the total event impact while satisfying all usage and grouping constraints.
\\

14 & 9 & 10 &
In a bustling city, a tech startup named InnovateTech is planning its annual hackathon. The event requires the allocation of resources across nine different project categories, each with varying levels of complexity and potential impact. The project categories are Algorithms, Machine Learning, Web Development, Mobile App Development, Data Visualization, Blockchain, Cybersecurity, IoT, and AI Ethics. The startup has a total of 8 slots available for project presentations. The sum of projects in Algorithms, Machine Learning, and Web Development cannot exceed 5; the sum of projects in Mobile App Development, Data Visualization, and Blockchain cannot exceed 3; and the sum of projects in Cybersecurity, IoT, and AI Ethics cannot exceed 2.

The allocation must also satisfy several coverage constraints. At least one project must be allocated across Algorithms, Mobile App Development, and Cybersecurity; at least one project must be allocated across Machine Learning, Data Visualization, and IoT; and at least one project must be allocated across Web Development, Blockchain, and AI Ethics. In addition, the total number of projects across Algorithms, Data Visualization, and AI Ethics cannot exceed 4; the total number of projects across Machine Learning, Mobile App Development, and Cybersecurity cannot exceed 3; and the total number of projects across Web Development, Data Visualization, and IoT cannot exceed 2.

The per-project costs for the nine categories are 2, 3, 4, 5, 6, 7, 8, 9, and 10 units, respectively. The startup wants to determine the optimal integer number of projects in each category to minimize total resource allocation cost while satisfying all slot, coverage, and grouping constraints.
 \\

\end{longtable}
\end{center}

\subsection{Example Solutions}
\label{app:example}

We take a problem from MAMO ComplexLP as an example and compare the solution generated by our trained LLM against that generated by the original Qwen3-8B-Base model.

Below we present the problem statement, the generation of \textsc{OPT-Zero}-8B, and the generation of Qwen3-8B-Base.
The trained model infers the correct integer-count formulation and returns the
optimal objective value $32$. The base model incorrectly restricts each food to
a binary inclusion decision and cannot produce feasible solution, which leads to a runtime error. These results indicate that despite the training without any external data, the LLM still learns how to use a correct variable type and finally solve the problem through the trial-and-error RL process.

\begin{tcolorbox}[
  enhanced,
  breakable,
  colback=white,
  colframe=black!35,
  boxrule=0.4pt,
  arc=1.5mm,
  left=1mm,
  right=1mm,
  top=1mm,
  bottom=1mm
]
\textbf{Problem Statement.}

Imagine you're meticulously planning your weekly meals to ensure you're hitting all your nutritional goals while sticking to a budget. You're considering five diverse food items, each with unique nutritional benefits and costs.

Here's a breakdown of what each food offers and its cost:

\begin{itemize} [leftmargin=*]
\item Food\_1: Delivers 3 grams of protein, 16 grams of carbohydrates, and 96 calories for \$9.
\item Food\_2: Provides 17 grams of protein, 24 grams of carbohydrates, and 182 calories for \$9.
\item Food\_3: Offers 16 grams of protein, 27 grams of carbohydrates, and 114 calories for just \$2.
\item Food\_4: Contains 8 grams of protein, 16 grams of carbohydrates, and 208 calories for \$9.
\item Food\_5: Supplies 6 grams of protein, 6 grams of carbohydrates, and 236 calories for \$5.
\end{itemize}

Your dietary targets for the week are to consume at least 100 grams of protein, 180 grams of carbohydrates, and 1796 calories. Your challenge is to determine the most cost-effective combination of these food items to meet or exceed these nutritional requirements. What is the cheapest cost to achieve your dietary goals?
\end{tcolorbox}

\begin{tcolorbox}[
  title={\textsc{OPT-Zero}-8B},
  colback=blue!2,
  colframe=blue!45!black,
  breakable
]

To solve this problem, we need to formulate an optimization model that minimizes
the total cost while meeting the nutritional requirements. Here's the
step-by-step reasoning:

\begin{enumerate} [leftmargin=*]
\item \textbf{Decision Variables}

Let \(x_1\) be the number of units of Food\_1.
Let \(x_2\) be the number of units of Food\_2.
Let \(x_3\) be the number of units of Food\_3.
Let \(x_4\) be the number of units of Food\_4.
Let \(x_5\) be the number of units of Food\_5.

All variables are non-negative integers since we cannot have a fraction of a
food item.

\item \textbf{Objective}

The goal is to minimize the total cost, which is given by:
\[
\text{Minimize} \quad 9x_1 + 9x_2 + 2x_3 + 9x_4 + 5x_5
\]

\item \textbf{Constraints}

\textbf{Protein Constraint}: The total protein must be at least 100 grams.
\[
3x_1 + 17x_2 + 16x_3 + 8x_4 + 6x_5 \geq 100
\]

\textbf{Carbohydrates Constraint}: The total carbohydrates must be at least
180 grams.
\[
16x_1 + 24x_2 + 27x_3 + 16x_4 + 6x_5 \geq 180
\]

\textbf{Calories Constraint}: The total calories must be at least 1796.
\[
96x_1 + 182x_2 + 114x_3 + 208x_4 + 236x_5 \geq 1796
\]

\item \textbf{Sanity Checks}

\textbf{Boundedness}: The objective function is bounded below by 0, and the
constraints ensure that the solution space is bounded.

\textbf{Feasibility}: There exists at least one feasible solution, such as
\[
x_1 = 10,\quad x_2 = 5,\quad x_3 = 10,\quad x_4 = 5,\quad x_5 = 10,
\]
which satisfies all constraints.

\item \textbf{Implementation Plan}

We will use Gurobi to implement the model, ensuring that the variable names and
constraints match the model formulation.
\end{enumerate}

\tcblower

\begin{lstlisting}
###family: MILP
###sense: min
###variables: x1:I:0:None; x2:I:0:None; x3:I:0:None; x4:I:0:None; x5:I:0:None
###objective: 9*x1 + 9*x2 + 2*x3 + 9*x4 + 5*x5
###constraints: 3*x1 + 17*x2 + 16*x3 + 8*x4 + 6*x5 >= 100;
                16*x1 + 24*x2 + 27*x3 + 16*x4 + 6*x5 >= 180;
                96*x1 + 182*x2 + 114*x3 + 208*x4 + 236*x5 >= 1796

###code:
from gurobipy import *

# 1) Build model
model = Model("Nutrition_Optimization")

# 2) Declare decision variables
x1 = model.addVar(vtype=GRB.INTEGER, lb=0, name="x1")
x2 = model.addVar(vtype=GRB.INTEGER, lb=0, name="x2")
x3 = model.addVar(vtype=GRB.INTEGER, lb=0, name="x3")
x4 = model.addVar(vtype=GRB.INTEGER, lb=0, name="x4")
x5 = model.addVar(vtype=GRB.INTEGER, lb=0, name="x5")

model.update()

# 3) Add constraints
model.addConstr(3*x1 + 17*x2 + 16*x3 + 8*x4 + 6*x5 >= 100, "Protein_Constraint")
model.addConstr(16*x1 + 24*x2 + 27*x3 + 16*x4 + 6*x5 >= 180, 
               "Carbohydrates_Constraint")
model.addConstr(96*x1 + 182*x2 + 114*x3 + 208*x4 + 236*x5 >= 1796,
                "Calories_Constraint")

model.update()

# 4) Set objective and optimize
model.setObjective(9*x1 + 9*x2 + 2*x3 + 9*x4 + 5*x5, GRB.MINIMIZE)
model.optimize()

# Extract and print the objective value
obj = model.ObjVal
print(f"objective_value={obj:.3f}")
###end
\end{lstlisting}
\end{tcolorbox}

\begin{tcolorbox}[
  title={\textsc{Qwen3-8B-Base}},
  colback=red!2,
  colframe=red!55!black,
  breakable
]

\begin{enumerate} [leftmargin=*]
\item Decision variables:

\begin{itemize}
\item x1: Binary (B), 0 <= x1 <= 1 (whether to include Food\_1)
\item x2: Binary (B), 0 <= x2 <= 1 (whether to include Food\_2)
\item x3: Binary (B), 0 <= x3 <= 1 (whether to include Food\_3)
\item x4: Binary (B), 0 <= x4 <= 1 (whether to include Food\_4)
\item x5: Binary (B), 0 <= x5 <= 1 (whether to include Food\_5)
\end{itemize}

\item Objective:

Minimize the total cost:
\[
9x_1 + 9x_2 + 2x_3 + 9x_4 + 5x_5
\]

\item Constraints:

C1: Protein constraint:
\[
3x_1 + 17x_2 + 16x_3 + 8x_4 + 6x_5 \geq 100
\]

C2: Carbohydrates constraint:
\[
16x_1 + 24x_2 + 27x_3 + 16x_4 + 6x_5 \geq 180
\]

C3: Calories constraint:
\[
96x_1 + 182x_2 + 114x_3 + 208x_4 + 236x_5 \geq 1796
\]

\item Sanity checks:

Boundedness: The objective cannot go to -infinity because all variables are
binary and bounded.

Feasibility: At least one feasible assignment exists, for example
\[
x_3 = 1,\quad x_5 = 1.
\]

\item Implementation plan:

Use the variable names x1, x2, x3, x4, x5.
Implement the objective and constraints as modeled.
\end{enumerate}

\tcblower

\begin{lstlisting}
###family: MILP
###sense: min
###variables: x1:B:0:1; x2:B:0:1; x3:B:0:1; x4:B:0:1; x5:B:0:1
###objective: 9*x1 + 9*x2 + 2*x3 + 9*x4 + 5*x5
###constraints: 3*x1 + 17*x2 + 16*x3 + 8*x4 + 6*x5 >= 100;
                16*x1 + 24*x2 + 27*x3 + 16*x4 + 6*x5 >= 180;
                96*x1 + 182*x2 + 114*x3 + 208*x4 + 236*x5 >= 1796

###code:
from gurobipy import *

# 1) Build model
model = Model("meal_planning")

# 2) Declare decision variables
x1 = model.addVar(vtype=GRB.BINARY, name="x1")
x2 = model.addVar(vtype=GRB.BINARY, name="x2")
x3 = model.addVar(vtype=GRB.BINARY, name="x3")
x4 = model.addVar(vtype=GRB.BINARY, name="x4")
x5 = model.addVar(vtype=GRB.BINARY, name="x5")

# 3) Add constraints
model.addConstr(3*x1 + 17*x2 + 16*x3 + 8*x4 + 6*x5 >= 100, "C1")
model.addConstr(16*x1 + 24*x2 + 27*x3 + 16*x4 + 6*x5 >= 180, "C2")
model.addConstr(96*x1 + 182*x2 + 114*x3 + 208*x4 + 236*x5 >= 1796, "C3")

# 4) Set objective and optimize
model.setObjective(9*x1 + 9*x2 + 2*x3 + 9*x4 + 5*x5, GRB.MINIMIZE)
model.optimize()

obj = model.ObjVal
print(f"objective_value={obj:.3f}")
###end
\end{lstlisting}
\end{tcolorbox}

\subsection{General-purpose Problem-solving Capability}
\label{app:general}

Beyond optimization modeling, we evaluate the trained model on GSM8K \cite{cobbe2021training}, GPQA-Diamond \cite{rein2024gpqa}, HumanEval \cite{chen2021evaluating}, and MBPP \cite{austin2021program}. These tasks assess mathematical reasoning, scientific question answering, and general code generation. We compare with the optimization-tuned StepORLM-8B \cite{zhou2026steporlm} and LLMOPT-14B \cite{jiang2025llmopt}, using the same input prompts and evaluation configurations for the compared models.

Table~\ref{tab:general-code-benchmarks} shows that \textsc{OPT-Zero}-8B achieves the highest score among the three models on all four tasks. It reaches 80.5\% HumanEval Pass@1 versus 18.9\% for StepORLM-8B, and 88.3\% GSM8K accuracy versus 20.9\% for LLMOPT-14B. These results demonstrate strong general reasoning and coding performance alongside optimization-modeling capability.

\begin{table}[thb]
\centering
\caption{General reasoning and coding performance of optimization-tuned models.}
\begin{tabular}{lcccc}
\toprule
Model & GSM8K Acc. & GPQA Acc. & HumanEval Pass@1 & MBPP Pass@1 \\
\midrule
StepORLM-8B & 82.1 & 33.8 & 18.9 & 47.4 \\
LLMOPT-14B & 20.9 & 39.4 & 50.6 & 65.8 \\
\textsc{OPT-Zero}-8B & \textbf{88.3} & \textbf{39.9} & \textbf{80.5} & \textbf{69.6} \\
\bottomrule
\end{tabular}
\label{tab:general-code-benchmarks}
\end{table}

Mathematical reasoning and code generation are central to interpreting multi-step optimization requirements and producing executable solutions. Performance on these external tasks complements the optimization benchmarks by assessing these skills in broader settings.

\begin{table}[t]
\footnotesize
\centering
\caption{Zero-shot solver transferability of different fine-tuned optimization modeling LLMs.}
\label{table:cross-solver}
\setlength{\tabcolsep}{3pt}
\begin{tabular}{lcccccc}
\toprule
Model & NL4Opt & EasyLP & ComplexLP & NLP4LP & IndOR & ReSocratic \\
\midrule
\multicolumn{7}{l}{\bf  CPLEX} \\
\quad ORLM-8B & 0.0 & 0.0 & 0.0 & 0.0 & 0.0 & 0.0 \\
\quad StepORLM-8B & 1.9 & 2.9 & 0.0 & 4.5 & 2.4 & 3.0 \\
\quad LLMOPT-14B & 23.5 & 36.7 & 27.9 & 39.3 & 19.1 & 26.8 \\
\quad \textsc{OPT-Zero}-8B & 92.5 & 94.5 & 34.2 & 94.9 & 57.1 & 80.1 \\
\midrule
\multicolumn{7}{l}{\bf  SCIP} \\
\quad ORLM-8B & 0.0 & 0.0 & 0.0 & 0.0 & 0.0 & 0.0 \\
\quad StepORLM-8B & 0.0 & 0.0 & 0.0 & 0.0 & 0.0 & 0.0 \\
\quad LLMOPT-14B & 22.1 & 13.6 & 9.0 & 16.3 & 11.9 & 15.9 \\
\quad \textsc{OPT-Zero}-8B & 92.0 & 96.1 & 55.9 & 97.2 & 47.6 & 80.9 \\
\midrule
\multicolumn{7}{l}{\bf  Pyomo} \\
\quad ORLM-8B & 0.0 & 0.0 & 0.0 & 0.0 & 0.0 & 0.0 \\
\quad StepORLM-8B & 2.4 & 0.0 & 0.0 & 2.8 & 0.0 & 0.7 \\
\quad \textsc{OPT-Zero}-8B & 90.1 & 93.6 & 59.5 & 94.4 & 47.6 & 78.2 \\
\bottomrule
\end{tabular}
\end{table}

\subsection{Zero-shot Solver Transferability}
\label{app:transfer}

In Figure~\ref{fig:solver_comparison_dumbbell}, we show that our self-play training framework can be used to teach models to work with different optimization solvers and modeling languages. More importantly, we empirically observe that \textsc{OPT-Zero} also demonstrates strong zero-shot transferability across solvers. Table~\ref{table:cross-solver} compares our method with several SFT(+Alignment)-based baselines on zero-shot generalization. Specifically, \textsc{OPT-Zero}-8B is trained on Gurobi, whereas ORLM-8B and StepORLM-8B are trained on the COPT solver; thus, evaluating them on CPLEX, SCIP, and Pyomo corresponds to an out-of-distribution setting. In contrast, LLMOPT-14B is trained on Pyomo, and is therefore included only in the comparisons on CPLEX and SCIP.

The Gurobi-trained \textsc{OPT-Zero}-8B model outperforms the listed baselines in every reported comparison across CPLEX, SCIP, and Pyomo (Table~\ref{table:cross-solver}). These results demonstrate strong zero-shot transfer of the trained weights to new solver interfaces. Together with the target-interface training experiments in Figure~\ref{fig:solver_comparison_dumbbell}, they show both the adaptability of the training framework and the transferability of the resulting model.

\subsection{Inference-time Scaling}
\label{app:scaling}

We explore how scaling inference-time techniques can improve reasoning capability of \textsc{OPT-Zero}. 

\begin{figure*}[t]
    \centering

    % Row 1
    \begin{minipage}[t]{0.32\textwidth}
        \centering
        \includegraphics[width=\textwidth]{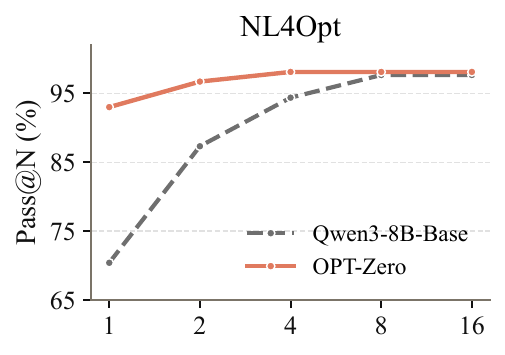}
    \end{minipage}
    \hfill
    \begin{minipage}[t]{0.32\textwidth}
        \centering
        \includegraphics[width=\textwidth]{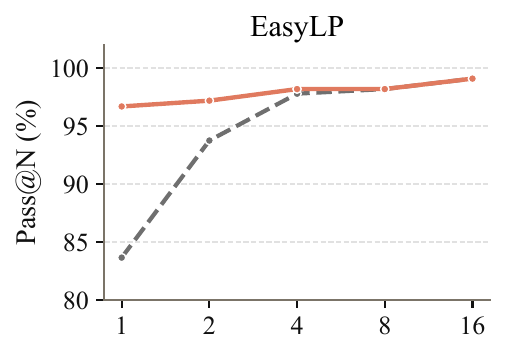}
    \end{minipage}
    \hfill
    \begin{minipage}[t]{0.32\textwidth}
        \centering
        \includegraphics[width=\textwidth]{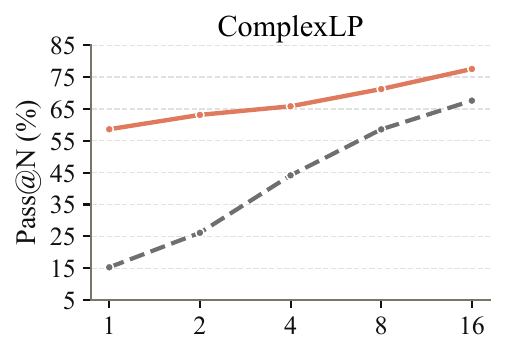}
    \end{minipage}

    \vspace{0.4em}

    % Row 2
    \begin{minipage}[t]{0.32\textwidth}
        \centering
        \includegraphics[width=\textwidth]{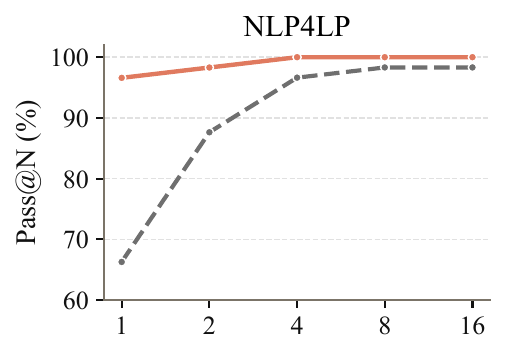}
    \end{minipage}
    \hfill
    \begin{minipage}[t]{0.32\textwidth}
        \centering
        \includegraphics[width=\textwidth]{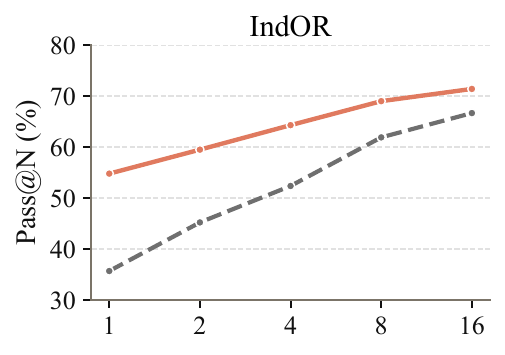}
    \end{minipage}
    \hfill
    \begin{minipage}[t]{0.32\textwidth}
        \centering
        \includegraphics[width=\textwidth]{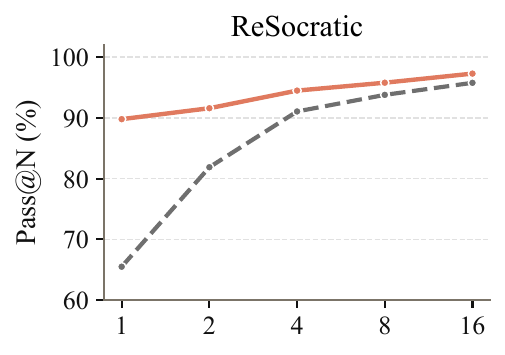}
    \end{minipage}

    \caption{Pass@$N$ (oracle coverage) across six benchmarks at equal candidate counts. An instance is covered if at least one of its $N$ sampled candidates matches the reference answer.}
    \label{fig:best_of_n_six}
\end{figure*}

\begin{figure}[thb]
    \centering
    \includegraphics[width=0.85\linewidth]{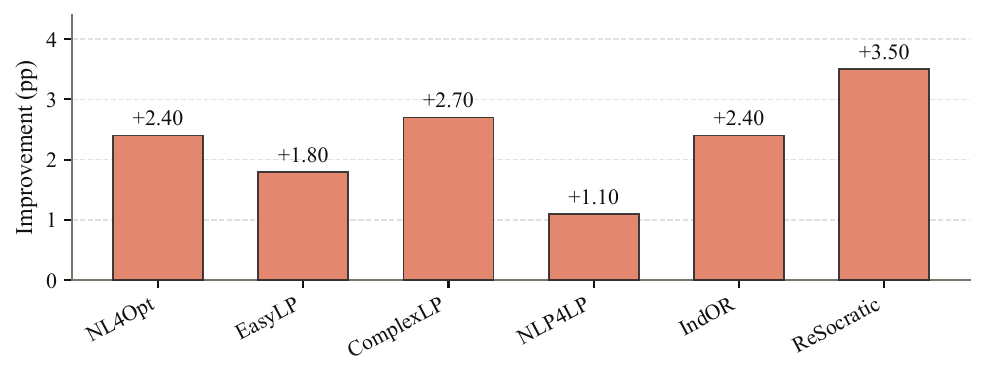}
    \caption{\textsc{OPT-Zero} performance improvement with the reflection mechanism.}
    \label{fig:reflection}
\end{figure}

\noindent \textbf{Pass@$N$ under equal candidate counts.} For each test instance, we sample $N\in\{1,2,4,8,16\}$ independent candidates and report oracle coverage: the fraction of instances with at least one candidate matching the reference answer. We compare \textsc{OPT-Zero} with Qwen3-8B-Base using the same candidate count at each $N$ (Figure~\ref{fig:best_of_n_six}).

\textsc{OPT-Zero} achieves higher coverage at small $N$ and retains an advantage on harder benchmarks such as IndustryOR, ComplexLP, and ReSocratic as $N$ increases. The gaps narrow on easier benchmarks as coverage approaches saturation. On NLP4LP, \textsc{OPT-Zero} reaches 100\% Pass@4, placing at least one correct solution among four candidates for every test problem. These results show that self-play training improves the quality of sampled solutions across inference budgets. Appendix~\ref{app:coe} further evaluates the combination with CoE inference.

\noindent \textbf{Self-Reflection on Execution Outcome.} In addition to independent candidate sampling, we can also use reflection mechanism for inference-time scaling: the Solver is able to revise its generation through the reflection on the executed outcome. We evaluate a one-round reflection variant of our optimization Solver at inference time. Starting from the standard solver prompt, the model first generates a single candidate solution, which is then executed and verified using the existing code-based evaluator. If the first attempt fails, we launch one additional reflection round that conditions on the original problem, the model’s previous code, a compact failure type, the verifier error message, and the previously computed objective value when available, while never exposing the correct formulation or objective to the Solver. According to the result shown in Figure~\ref{fig:reflection}, we can find reflection consistently improves performance. This suggests that a single feedback-driven revision step can meaningfully improve optimization code generation, demonstrating that \textsc{OPT-Zero} can achieve better performance with appropriate inference-time reasoning techniques.

\subsection{Effect of Difficulty-Aware Sampling}
\label{app:difficulty}

\begin{figure}[htb]
    \centering
    \includegraphics[width=0.95\linewidth]{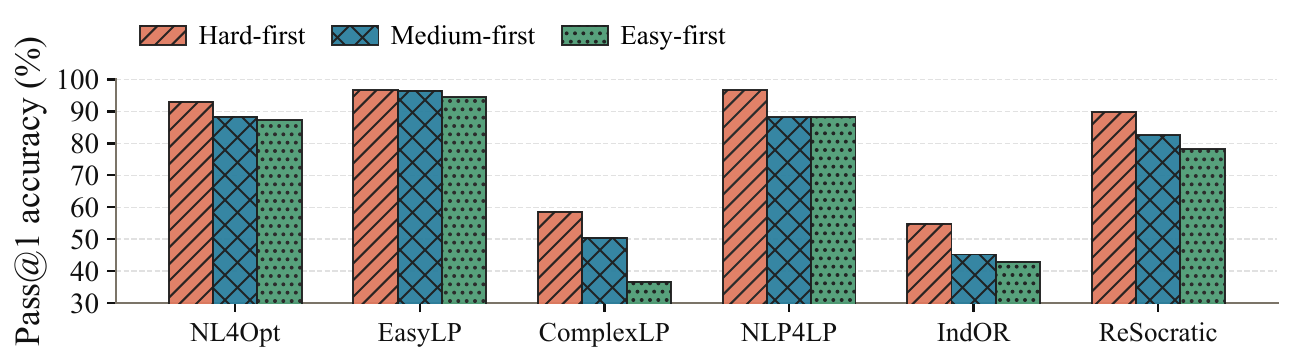}
    \caption{Model performance with different sampling strategies.}
    \label{fig:difficulty}
\end{figure}

During Solver training, we adopt difficulty-aware sampling with weights $(0.2, 0.3, 0.5)$ for easy, medium, and hard problems, respectively, thereby prioritizing harder instances. This hard-instance-first strategy yields denser and more informative learning signals, since hard problems are more likely to reveal the model’s true failure modes instead of repeatedly rewarding behaviors that have already been learned from easy cases. To assess the effectiveness of this strategy, we further compare it against an easy-instance-first setting and a medium-instance-first setting, with sampling weights of $(0.5, 0.3, 0.2)$ and $(0.3, 0.5, 0.2)$, respectively.

The comparative results of the three strategies are shown in Figure~\ref{fig:difficulty}. It can be found that sampling more hard instances leads to the best performance. When easier instances are prioritized during training, model performance shows a more significant decline on relatively difficult benchmarks (e.g., ComplexLP and IndOR). This indicates that learning on more difficult problems can also enhance the LLM in fulfilling more complex optimization modeling tasks. As a result, pre-scoring for the Proposer generations (at stage 2) and difficulty refresh (at stage 4) play important roles in the training of \textsc{OPT-Zero} because they provide reliable labels when choosing the training set of the Solver.

\subsection{Comparison to Training with Solve-rate Reward}
\label{app:dpo}

\begin{table}[thb]
\centering
\small
\caption{Comparison between our method and the solve-rate-based DPO baseline.}
\label{tab:solve-rate-baseline}
\begin{tabular}{lcccccc}
\toprule
Method & NL4Opt & EasyLP & ComplexLP & NLP4LP & IndOR & ReSocratic \\
\midrule
Solve-rate-based DPO & 86.8 & 89.4 & 29.7 & 88.8 & 42.9 & 80.9 \\
Structural-reward GRPO & \textbf{93.0} & \textbf{96.7} & \textbf{58.6}& \textbf{96.6} & \textbf{54.8} & \textbf{89.8} \\
\bottomrule
\end{tabular}
\end{table}

For general coding tasks, prior work has used the solve rate over multiple Solver generations as the reward for the Proposer \cite{zhao2025absolute}. The underlying intuition is that generations with lower solve rates are more challenging for the Solver, and can therefore provide more effective training signals. However, using raw solve rate as the Proposer’s GRPO reward would require multiple online solver rollouts for each proposed instance to obtain a reliable estimate, which makes training substantially more expensive. In our implementation, we design fine-grained Proposer reward and avoid calculating online multiple Solver generations during Proposer training. 

To examine an alternative use of solve-rate supervision, we construct an offline solve-rate-based baseline using Direct Preference Optimization (DPO). Specifically, we first sample multiple Proposer completions, estimate their solve rates with a fixed number of Solver rollouts, and then build preference pairs from these offline estimates. This baseline preserves the core intuition of solve-rate supervision while avoiding repeated online Solver sampling inside Proposer training. In our experiments, the DPO baseline uses 500 Proposer prompts, 8 Proposer samples per prompt, and 8 Solver rollouts per valid instance for solve-rate estimation. We construct up to 500 preference pairs per iteration, where chosen samples have solve rates in the range $[0.25, 0.75]$, while other samples are regarded as rejected samples. All other components, including the Proposer backbone, Solver backbone, and evaluation protocol, are kept the same as in our main method.

Table~\ref{tab:solve-rate-baseline} shows that structural-reward GRPO outperforms solve-rate-based DPO on all six benchmarks, with a particularly large gain on ComplexLP (58.6\% versus 29.7\%). This supports our training design, which combines optimization-specific structural feedback with GRPO and obtains Proposer rewards directly from formulation and code verification.

\subsection{Fine-tuning Different LLMs}
\label{app:llms}

\definecolor{gainblue}{RGB}{30,100,200}
\newcommand{\gain}[1]{\textcolor{gainblue}{\scriptsize $+#1$}}

\begin{table}[ht]
\centering
\caption{Performance of \textsc{OPT-Zero} in fine-tuning LLMs with different architectures and sizes.}
\label{tab:diff_llms}
\setlength{\tabcolsep}{4pt}
\renewcommand{\arraystretch}{1.12}
\resizebox{\textwidth}{!}{%
\begin{tabular}{lcccccccc}
\toprule
\textbf{Model}
& \textbf{NL4Opt}
& \textbf{EasyLP}
& \textbf{ComplexLP}
& \textbf{NLP4LP}
& \textbf{IndOR}
& \textbf{ReSocratic}
& \textbf{Macro Avg.}
& \textbf{Micro Avg.} \\
\midrule

\textsc{Qwen3-8B-Base}
& 70.4 & 83.7 & 15.3 & 66.3 & 35.7 & 65.5 & 56.2 & 68.4 \\
\quad +\textsc{OPT-Zero}
& 93.0 & 96.7 & 58.6 & 96.6 & 54.8 & 89.8 & 81.6 & 90.3 \\
\quad \textcolor{gainblue}{Absolute Gain ($\Delta$)}
& \gain{22.6} & \gain{13.0} & \gain{43.3} & \gain{30.3}
& \gain{19.1} & \gain{24.3} & \gain{25.4} & \gain{21.9} \\

\addlinespace[2pt]
\textsc{Qwen3.5-4B}
& 12.7 & 24.4 & 14.4 & 11.2 & 7.1 & 12.7 & 13.8 & 16.8 \\
\quad +\textsc{OPT-Zero}
& 91.6 & 95.2 & 51.4 & 96.1 & 57.1 & 93.1 & 80.8 & 89.9 \\
\quad \textcolor{gainblue}{Absolute Gain ($\Delta$)}
& \gain{78.9} & \gain{70.8} & \gain{37.0} & \gain{84.9}
& \gain{50.0} & \gain{80.4} & \gain{67.0} & \gain{73.1} \\

\addlinespace[2pt]
\textsc{Granite-4.1-8B}
& 51.2 & 38.5 & 29.7 & 50.6 & 14.3 & 54.3 & 39.8 & 44.7 \\
\quad +\textsc{OPT-Zero}
& 80.8 & 97.8 & 30.6 & 81.5 & 28.6 & 72.7 & 65.3 & 79.7 \\
\quad \textcolor{gainblue}{Absolute Gain ($\Delta$)}
& \gain{29.6} & \gain{59.3} & \gain{0.9} & \gain{30.9}
& \gain{14.3} & \gain{18.4} & \gain{25.5} & \gain{35.0} \\

\addlinespace[2pt]
\textsc{Gemma-4-31B-it}
& 76.1 & 87.2 & 60.4 & 84.8 & 52.4 & 79.9 & 73.5 & 80.4 \\
\quad +\textsc{OPT-Zero}
& 93.0 & 93.2 & 82.9 & 97.2 & 73.8 & 94.8 & 89.1 & 92.8 \\
\quad \textcolor{gainblue}{Absolute Gain ($\Delta$)}
& \gain{16.9} & \gain{6.0} & \gain{22.5} & \gain{12.4}
& \gain{21.4} & \gain{14.9} & \gain{15.6} & \gain{12.4} \\

\bottomrule
\end{tabular}
}
\end{table}

We evaluate the generality of \textsc{OPT-Zero} by applying the same self-play framework to Qwen3-8B-Base, Qwen3.5-4B, IBM Granite-4.1-8B, and Google Gemma4-31B-IT. For Gemma4-31B-IT, we use rank-128 LoRA with approximately 0.98B trainable parameters. Table~\ref{tab:diff_llms} reports the before/after scores on all six benchmarks.

The results show improvements for every backbone. Qwen3-8B-Base improves from 56.2/68.4 to 81.6/90.3 Macro/Micro, Qwen3.5-4B from 13.8/16.8 to 80.8/89.9, and Granite-4.1-8B from 39.8/44.7 to 65.3/79.7. Gemma4-31B-IT improves from 73.5/80.4 to 89.1/92.8, gaining 15.6 Macro and 12.4 Micro percentage points from a strong instruction-tuned initialization. The framework therefore supports both full-parameter training and parameter-efficient adaptation across different model families.

Together, these results demonstrate the broad applicability of \textsc{OPT-Zero} across model architectures, sizes, and initializations. Appendix~\ref{app:challenging} extends the Gemma4-31B-IT evaluation to OptMATH-Bench and MIPLIB-NL, while Appendix~\ref{app:positioning} discusses combining self-play with external training data.

\subsection{Transfer to Challenging External Benchmarks}
\label{app:challenging}

\paragraph{Benchmarks and evaluation setting.}
We evaluate Gemma4-31B-IT before and after \textsc{OPT-Zero} on two additional held-out benchmarks: OptMATH-Bench \cite{lu2025optmath} and MIPLIB-NL \cite{li2026constructing}. OptMATH-Bench provides a further test of optimization modeling, while MIPLIB-NL evaluates natural-language modeling of industrial-scale mixed-integer linear programs constructed from MIPLIB 2017. These benchmarks assess transfer beyond the six-benchmark suite to tasks with greater scale and structural complexity.

\begin{table}[htbp]
\centering
\small
\caption{Gemma4-31B-IT performance before and after \textsc{OPT-Zero} on two additional held-out benchmarks. Scores are objective-value-based Pass@1 (\%); gains are in percentage points.}
\label{tab:challenging_benchmarks}
\begin{tabular}{lrr}
\toprule
Model & OptMATH-Bench & MIPLIB-NL\\
\midrule
Gemma4-31B-IT & 41.0 & 8.92 \\
\quad +\textsc{OPT-Zero} & \textbf{48.2} & \textbf{11.74} \\
\midrule
Absolute gain (percentage points) & +7.2 & +2.82 \\
\bottomrule
\end{tabular}
\end{table}

\paragraph{Single-generation evaluation.}
Both models use the single-generation configuration in Appendix~\ref{app:setting}: temperature 0.1, top-$p$ 0.95, at most 8192 new tokens, a 60-second execution limit, and objective matching with 5\% relative tolerance and $10^{-5}$ absolute tolerance for near-zero objectives. Evaluation covers all 166 OptMATH-Bench instances and the 213 MIPLIB-NL instances with numerical reference objectives from the 223-instance release. Both models use these same evaluation populations.

\paragraph{Results.}
\textsc{OPT-Zero} improves accuracy from 41.0\% to 48.2\% on OptMATH-Bench and from 8.92\% to 11.74\% on MIPLIB-NL (Table~\ref{tab:challenging_benchmarks}). The improvements demonstrate that self-play training benefits a strong instruction-tuned model on externally authored optimization problems, including industrial-scale MILP tasks.

\subsection{Empirical Difficulty of the Self-generated Curriculum}
\label{app:curriculum_analysis}

\paragraph{Candidate collection and measurement.}
We analyze all canonically verified Proposer problems collected over 15 self-play iterations, before solve-rate filtering and buffer admission. Each candidate $p$ generated at iteration $t$ is evaluated with $H=8$ rollouts of the contemporaneous Solver. Its empirical solve rate is $\hat{\rho}_t(p)=k_t(p)/8$, where $k_t(p)$ counts successful executions under the training-time objective-matching criterion in Section~\ref{sec:loop}. Table~\ref{tab:curriculum_success} pools these measurements across all 15 iterations, characterizing task difficulty relative to the Solver as training progresses.

\begin{table}[htbp]
\centering
\small
\caption{Empirical solve-rate distribution of canonically verified Proposer candidates across 15 iterations. Each candidate is evaluated with eight rollouts of the contemporaneous Solver before filtering. Buffer admission further applies structural deduplication and capacity control.}
\label{tab:curriculum_success}
\begin{tabularx}{\textwidth}{l r X X}
\toprule
Successful rollouts & Share & Empirical interpretation & Treatment at the solve-rate gate \\
\midrule
$0/8$ & 55.7\% & No successful rollout observed & Excluded at this scoring step \\
$1$--$6/8$ & 32.3\% & Mixed outcomes; hard or medium & Eligible for retention \\
$7$--$8/8$ & 12.0\% & High observed solve rate & $7/8$ eligible (easy); $8/8$ excluded \\
\bottomrule
\end{tabularx}
\end{table}

\paragraph{Complementary generation and reconstruction tasks.}
The Proposer jointly generates a formulation, implementation, and story. The Solver receives only the story and reconstructs a formulation and executable implementation, creating a distinct learning task within the shared model. The Solver fails in all eight attempts on 55.7\% of verified candidates, while 32.3\% produce both successful and unsuccessful attempts. These mixed-outcome candidates supply learning opportunities with observable success and failure. Solver updates use externally computed rewards from formulation and code verification.

\paragraph{From generated candidates to training samples.}
The solve-rate filter retains candidates with $0<\hat{\rho}<1$. With eight rollouts, hard problems have $k=1,2$, medium problems have $k=3,4,5,6$, and easy problems have $k=7$. Candidates with $k=0$ or $k=8$ are excluded at this step; the 12.0\% high-success bin therefore contains both retained $7/8$ cases and excluded $8/8$ cases. Structural deduplication, capacity control, and replay maintain the reference buffer. Solver training then samples easy, medium, and hard problems with weights $(0.2,0.3,0.5)$, respectively. Appendix~\ref{app:difficulty} evaluates alternative sampling policies.

\paragraph{Relation to the learning signal.}
The curriculum selects problems that provide useful feedback for the current Solver: filtering $k=0$ favors candidates with an observed successful trajectory, while filtering $k=8$ directs training toward problems the Solver has yet to master consistently. This problem-level selection complements DAPO-style rollout-group filtering by total-reward variance. The Solver's total reward includes partial credit for formulation and code execution, so even unsuccessful rollouts can receive different rewards. GRPO learns from within-group reward differences, while Proposer updates, the reference buffer, and adaptive structural targets continually shape the tasks presented to the Solver.

\paragraph{Connection to external task performance.}
The component ablations in Table~\ref{table:ablation} connect curriculum construction to held-out performance. Removing Proposer training reduces Macro/Micro accuracy from 81.6/90.3 to 75.1/84.3, and removing the reference buffer reduces it to 72.9/83.2. Together with the candidate statistics, these results show how Proposer training and buffer-based task reuse contribute to learning and external task performance.

\subsection{Combining Self-play Training with Chain-of-Experts Inference}
\label{app:coe}

\paragraph{Experimental setup.}
We apply Chain-of-Experts (CoE) \cite{xiao2023chain} to both Qwen3-8B-Base and \textsc{OPT-Zero}-8B to evaluate the combination of self-play training and inference-time scaling. Single-generation inference uses one LLM call per problem, while CoE uses approximately eight calls with the same procedure for both models. Table~\ref{tab:coe_controlled} reports all six benchmarks and their Macro/Micro averages. The single-generation baseline is taken from Table~\ref{table:main_results}.

\begin{table}[htbp]
\centering
\scriptsize
\setlength{\tabcolsep}{2.5pt}
\renewcommand{\arraystretch}{1.18}
\caption{Single-generation and CoE results on the six-benchmark suite (\%). Single-generation uses one LLM call per problem, while CoE uses approximately eight calls with the same procedure for both models.}
\label{tab:coe_controlled}
\begin{tabularx}{\textwidth}{l*{8}{>{\centering\arraybackslash}X}}
\toprule
Model / inference & NL4Opt & EasyLP & ComplexLP & NLP4LP & IndOR & ReSocratic & Macro & Micro \\
\midrule
Qwen3-8B-Base / single & 70.4 & 83.7 & 15.3 & 66.3 & 35.7 & 65.5 & 56.2 & 68.4 \\
Qwen3-8B-Base / CoE & 87.3 & 83.5 & 45.0 & 87.6 & 40.5 & 83.4 & 71.2 & 80.4 \\
\midrule
\textsc{OPT-Zero}-8B / single & 93.0 & 96.7 & 58.6 & 96.6 & \textbf{54.8} & 89.8 & 81.6 & 90.3 \\
\textsc{OPT-Zero}-8B / CoE & \textbf{95.4} & \textbf{96.7} & \textbf{62.2} & \textbf{96.6} & 52.4 & \textbf{93.3} & \textbf{82.8} & \textbf{91.8} \\
\bottomrule
\end{tabularx}
\end{table}

\paragraph{Complementary gains from training and inference.}
CoE improves Qwen3-8B-Base from 59.9 to 71.2 Macro and from 71.5 to 80.4 Micro. Single-generation \textsc{OPT-Zero}-8B reaches 81.6 Macro and 90.3 Micro, outperforming the CoE-enhanced baseline by 10.4 and 9.9 percentage points, respectively, with one LLM call per problem.

Applying CoE to \textsc{OPT-Zero}-8B further increases Macro accuracy to 82.8 and Micro accuracy to 91.8. Under the same CoE procedure, the trained model leads Qwen3-8B-Base by 11.6 Macro and 11.4 Micro percentage points. These results demonstrate that self-play training and inference-time scaling are complementary: training strengthens the model, and CoE further improves its aggregate performance.

% AUTHOR CHECK: Base identity is confirmed. Archive the exact checkpoint, prompt/template and concrete CoE implementation/configuration for the approximately eight-call protocol.

\subsection{Contribution of Proposer Code Generation}
\label{app:proposer_code}

\paragraph{Learning executable implementations.}
The canonical verifier translates a declared formulation into a solver model and obtains its reference objective. The language model learns to implement that formulation through the requested solver interface, including API calls, variable indexing, data handling, status checks, and objective extraction. Canonical solving thus supplies objective supervision, while code generation develops the implementation capability needed for deployment.

The Proposer generates code for its own formulation, adding formulation-to-implementation training alongside the Solver's description-to-formulation-and-code task. Both roles update the same model. Canonical validity, Proposer code consistency, and the Solver's model/code rewards provide feedback for these complementary tasks (Sections~\ref{sec:proposer_reward} and~\ref{sec:solver_reward}).

\paragraph{Ablation and result.}
We remove both the Proposer's code output and its code-correctness reward factor. The Proposer continues to generate the formulation and natural-language story, with reward $R^P_{\mathrm{no\mbox{-}code}}=R_{\mathrm{valid}}R_{\mathrm{struct}}$. The Solver retains its formulation and executable-code outputs and its original reward; all other training settings and the six-benchmark evaluation protocol remain unchanged. Micro-average accuracy decreases from 90.3\% to 87.1\%, a reduction of 3.2 percentage points (Table~\ref{tab:proposer_code_ablation}).

\begin{table}[htbp]
\centering
\small
\caption{Ablation of Proposer code generation and its code-correctness reward. The Solver retains its original outputs and reward. Scores are Micro-average objective-value accuracy (\%).}
\label{tab:proposer_code_ablation}
\begin{tabular}{lr}
\toprule
Training configuration & Micro Avg. (\%) \\
\midrule
Full \textsc{OPT-Zero}-8B & \textbf{90.3} \\
Without Proposer code generation & 87.1 \\
\midrule
Difference (full minus ablated, percentage points) & +3.2 \\
\bottomrule
\end{tabular}
\end{table}

The 3.2-point gap demonstrates the contribution of the Proposer code branch to end-to-end modeling. With shared parameters, this branch adds practice in translating formulations into executable implementations, complemented by code-correctness feedback. Together with the component ablations in Table~\ref{table:ablation}, the result supports jointly training formulation and implementation skills.

\subsection{Textual Diversity of Generated Training Problems}
\label{app:text_diversity}

\paragraph{Sample construction.}
We compare 300 retained \textsc{OPT-Zero} reference-buffer problems sampled across self-play iterations with 300 OptMATH-Train problems \cite{lu2025optmath}. The OptMATH sample is drawn from eight spread-out regions of the corpus. We analyze the natural-language statements in both samples using text-similarity metrics to quantify diversity and repetition.

\paragraph{Representation and metrics.}
We use word unigram and bigram TF-IDF features with cosine similarity. Let $x_i$ denote the normalized feature vector of statement $i$ and $K_{ij}=x_i^\top x_j$ its within-corpus similarity matrix, with $K_{ii}=1$. For $n=300$, let $\lambda_1,\ldots,\lambda_n$ be the eigenvalues of $K/n$. The Vendi Score \cite{friedman2023vendi} is
\begin{equation}
\operatorname{VS}(K)=\exp\!\left(-\sum_{i=1}^{n}\lambda_i\log\lambda_i\right),
\qquad 0\log0:=0.
\label{eq:text_vendi}
\end{equation}
It expresses the effective number of distinct samples under the chosen similarity function, ranging from 1 for identical representations to $n$ for mutually orthogonal representations. We also report nearest-neighbor redundancy,
\begin{equation}
\operatorname{NN}(K)=\frac{1}{n}\sum_{i=1}^{n}\max_{j\ne i}K_{ij},
\label{eq:text_nn_redundancy}
\end{equation}
which averages each statement's similarity to its closest other statement. Higher VS and lower NN redundancy indicate greater diversity under this text representation.

\begin{table}[htbp]
\centering
\small
\caption{Textual diversity of two samples of 300 problem statements, measured using word 1--2 gram TF-IDF features and cosine similarity.}
\label{tab:text_diversity}
\begin{tabular}{lrrr}
\toprule
Corpus sample & Statements & Vendi Score $\uparrow$ & NN redundancy $\downarrow$ \\
\midrule
\textsc{OPT-Zero} reference buffer & 300 & \textbf{244} & \textbf{0.32} \\
OptMATH-Train & 300 & 178 & 0.56 \\
\bottomrule
\end{tabular}
\end{table}

\paragraph{Results.}
The \textsc{OPT-Zero} buffer sample achieves a higher Vendi Score (244 versus 178) and lower nearest-neighbor redundancy (0.32 versus 0.56) than the OptMATH-Train sample. These results show greater textual diversity and less repetitive wording in the sampled self-play problems. Vendi summarizes diversity across the full similarity spectrum, while nearest-neighbor redundancy captures similarity to the closest other problem statement.

This analysis complements the structural rewards and buffer deduplication described in Appendix~\ref{app:structural_reward}, as well as the empirical task-difficulty analysis in Appendix~\ref{app:curriculum_analysis}. Together, these components and measurements characterize how \textsc{OPT-Zero} constructs a varied, adaptive training curriculum.
% Reproducibility metadata: archive sampled instance IDs, buffer iterations, the eight OptMATH sampling regions, random seed, and TF-IDF vocabulary/IDF fitting configuration.

\subsection{Adaptive Supervision and Relation to Prior Work}
\label{app:positioning}

\paragraph{Research question.}
\textsc{OPT-Zero} studies optimization-modeling post-training through a solver-verifiable curriculum that evolves with the learner. The model generates its own training instances and receives feedback from optimization solvers. This approach complements reusable synthetic corpora by continually generating and selecting tasks as the model's capabilities change.

\paragraph{Relation to corpus-based training and prior self-play.}
Table~\ref{tab:supervision_positioning} compares task sources, learned roles, and feedback mechanisms. OptMATH supplies a synthesized training corpus; SIRL applies solver feedback to externally constructed problems; and StepORLM co-evolves a policy and a process verifier using a synthesized training set. Absolute Zero and R-Zero develop task-generating self-play for broader reasoning settings. \textsc{OPT-Zero} brings this learning principle to end-to-end optimization modeling by combining canonical formulation solving, executable-code feedback, optimization-specific structural rewards, and a difficulty-aware reference buffer.

\begin{table}[htbp]
\centering
\footnotesize
\setlength{\tabcolsep}{4pt}
\renewcommand{\arraystretch}{1.15}
\caption{Task sources, learned roles, and feedback mechanisms in the primary configurations of representative methods.}
\label{tab:supervision_positioning}
\begin{tabularx}{\textwidth}{>{\raggedright\arraybackslash}p{0.17\textwidth} *{3}{>{\raggedright\arraybackslash}X}}
\toprule
Method & Task source and adaptation & Learned roles & Feedback and scope \\
\midrule
OptMATH \cite{lu2025optmath} & Seed formulations and generators support bidirectional synthesis of a training corpus & Downstream optimization model trained on synthesized examples & Forward reconstruction and validation in corpus construction \\
\addlinespace
SIRL \cite{chen2025solverinformed} & Externally constructed, filtered optimization tasks & Optimization-modeling policy updated by RL & Solver execution, objective accuracy, and instance-level modeling feedback \\
\addlinespace
StepORLM \cite{zhou2026steporlm} & Teacher-synthesized training set; new solution trajectories collected during co-evolution & Policy and generative process reward model & Solver outcome feedback and learned process supervision \\
\addlinespace
Absolute Zero \cite{zhao2025absolute} & Self-generated code-reasoning tasks; no external task corpus & A shared model proposes and solves tasks & Code execution validates tasks and answers \\
\addlinespace
R-Zero \cite{huang2026rzero} & Challenger-generated reasoning tasks; no external task corpus & Separately optimized Challenger and Solver in the primary configuration & Solver-relative challenge and solution feedback \\
\addlinespace
\textsc{OPT-Zero} & New optimization problems generated and pre-scored during training; difficulty-refreshed buffer & One shared model alternates Proposer and Solver updates & Canonical formulation solving, code checks, and adaptive structural targets \\
\bottomrule
\end{tabularx}
\end{table}

\paragraph{Compatibility with external data.}
The main \textsc{OPT-Zero} configuration generates its optimization training problems through self-play. The data-augmented variant, \textsc{OPT-Zero}-D, draws 30\% of Solver training problems per iteration from the training set of \citet{chen2025solverinformed}, together with their reference objectives. Its results in Table~\ref{table:main_results} demonstrate the compatibility of self-generated supervision with external training data.

\paragraph{Solver-independent objectives and interface-specific implementation.}
Reference objectives provide a common supervision signal across solver interfaces. Generating solutions with DOcplex, PySCIPOpt, or Pyomo also requires expressing variables, constraints, data access, solve calls, and result extraction through each API. Figure~\ref{fig:solver_comparison_dumbbell} evaluates training with different target interfaces, while Appendix~\ref{app:transfer} evaluates Gurobi-trained weights directly on new interfaces. Together, these experiments assess framework adaptability and zero-shot model transfer.

\paragraph{Adaptation across starting models.}
Table~\ref{tab:diff_llms} shows improvements across four backbones. Qwen3.5-4B increases from 16.8\% to 89.9\% Micro-average, while Gemma4-31B-IT increases from 80.4\% to 92.8\% using rank-128 LoRA. Self-play therefore benefits both base models and a strong instruction-tuned model, using full-parameter or parameter-efficient training. Appendix~\ref{app:challenging} further demonstrates improvements on harder external benchmarks.

\section{Frequently Asked Questions}

\paragraph{Q1: What exactly does data-free mean in \textsc{OPT-Zero}?}

In the main configuration, \textsc{OPT-Zero} generates its optimization post-training instances through self-play. It uses no external optimization training corpus, human-annotated formulations, expert-written solution programs, or teacher-generated solutions. A pretrained model supplies the starting policy, and solver-grounded interaction produces the subsequent training supervision.

Prompts, canonical parsers, solver interfaces, reward functions, and curriculum rules define the training environment. They specify how tasks are represented, verified, and selected. Appendix~\ref{app:compute_cost} reports the computational cost of task generation, verification, and training.

The \textsc{OPT-Zero}-D variant combines self-generated problems with external training data. Appendix~\ref{app:positioning} compares these supervision sources and the optimization-specific design with related self-play methods.

\paragraph{Q2: Why focus on LP, IP, and MILP?}

LP, IP, and MILP are widely used in optimization-modeling benchmarks and prior training-based methods, providing a common setting for evaluating \textsc{OPT-Zero}.

These problem classes also offer mature exact solvers and reliable feasibility, boundedness, and objective-value checks. Generated formulations can therefore be parsed, solved, and evaluated automatically, providing executable feedback for self-play training.

LP/IP/MILP covers practical decision problems in allocation, scheduling, production planning, assignment, selection, and routing. Integer and mixed-integer formulations introduce discrete decisions and combinatorial structure, making this a substantive test of optimization-modeling capability.

Extending solver-grounded self-play to nonlinear, stochastic, robust, and multi-objective optimization is a promising direction. Such extensions require suitable problem representations and reliable solver feedback.

\paragraph{Q3: Since the Proposer generates both the natural-language story and the formulation, how do we know the story is consistent with the mathematical model?}
\label{app:semantic_scope}

The pipeline combines canonical formulation solving, code-execution feedback, and story-based reconstruction. Canonical verification independently solves the declared formulation to obtain its reference objective, while code verification checks the generated implementation against that objective. Story--formulation alignment is encouraged through the Proposer prompt and the Solver's reconstruction task.

The Proposer prompt requires the story to preserve the formulation's variables, coefficients, bounds, objective, and constraints. The Solver receives only the story and independently reconstructs a formulation and executable code. Its empirical solve rate guides curriculum selection, and the buffer retains candidates with mixed outcomes as specified in Section~\ref{sec:loop}. Appendix~\ref{app:curriculum_analysis} reports these solve rates throughout training.

Invalid canonical formulations receive zero validity reward. For valid formulations, incorrect Proposer code receives the reduced correctness factor $\alpha_c>0$ (Appendix~\ref{app:structural_reward}). Solve-rate filtering and structural deduplication then select problems for the reference buffer. These feedback mechanisms jointly shape the generated curriculum. We evaluate the resulting Solver on independently authored benchmark descriptions and reference answers, including OptMATH-Bench and MIPLIB-NL (Appendix~\ref{app:challenging}). The gains on these external tasks demonstrate the usefulness of self-generated training problems for optimization modeling.

\paragraph{Q4: How do hand-designed rules support self-play?}

Optimization modeling has a formal structure that supports algorithmic checks of variable domains, constraints, feasibility, boundedness, and optimality. \textsc{OPT-Zero} uses this structure to construct rewards from formulation and code verification. The rules define the feedback environment, while the model learns to generate problems and solutions through interaction with it. The ablations in Table~\ref{table:ablation} demonstrate the contribution of these components: removing Solver training, Proposer training, or the reference buffer, or replacing structural rewards with binary rewards, reduces performance. Together, the verifier, structural rewards, and reference buffer support effective self-play learning.

\end{document}